\RequirePackage{fix-cm}
\documentclass[twocolumn]{svjour3}          
\smartqed  
\usepackage{graphicx}
\usepackage{natbib}
\usepackage{amsmath,amssymb}
\usepackage[nolist]{acronym}
\usepackage{color}
\usepackage{tikz}
\usepackage{tabularx}
\usepackage{appendix}
\usepackage{units}
\usepackage{algorithm}
\usepackage{algpseudocode} 
\usepackage{pifont}
\usepackage{pgfplots}
\usepackage{xparse}

\newcommand{\approptoinn}[2]{\mathrel{\vcenter{
			\offinterlineskip\halign{\hfil$##$\cr
				#1\propto\cr\noalign{\kern2pt}#1\sim\cr\noalign{\kern-2pt}}}}}

\newcommand{\cmark}{\ding{51}}%
\newcommand{\xmark}{\ding{55}}%
\NewDocumentCommand{\rot}{O{45} O{0.6cm} m}{\makebox[#2][l]{\rotatebox{#1}{#3}}}%
\renewcommand{\d}[1]{\ensuremath{\operatorname{d}\!{#1}}}
\DeclareMathOperator*{\sgn}{sgn}

\DeclareMathOperator*{\trace}{trace}
\DeclareMathOperator*{\argmax}{arg\,max}
\definecolor{dlrred}{RGB}{179,63,61}
\definecolor{dlrblue}{RGB}{0,117,187}
\journalname{International Journal of Computer Vision}
\begin{document}

\title{SRT3D: A Sparse Region-Based 3D Object Tracking Approach for the Real World
}


\author{Manuel Stoiber \and Martin Pfanne \and Klaus H. Strobl \and Rudolph Triebel \and Alin Albu-Sch\"affer
}


\institute{M. Stoiber \at
           German Aerospace Center, 82234 Wessling, Germany \\
           Technical University of Munich, 80333 Munich, Germany\\
           \email{manuel.stoiber@dlr.de}
           \and
           M. Pfanne \at
           German Aerospace Center, 82234 Wessling, Germany \\
           \email{martin.pfanne@dlr.de}
           \and
           K. H. Strobl \at
           German Aerospace Center, 82234 Wessling, Germany \\
           \email{klaus.strobl@dlr.de}
           \and
           R. Triebel \at
           German Aerospace Center, 82234 Wessling, Germany \\
           Technical University of Munich, 80333 Munich, Germany\\
           \email{rudolph.triebel@dlr.de}
           \and
           A. Albu-Sch\"affer \at
           German Aerospace Center, 82234 Wessling, Germany \\
           Technical University of Munich, 80333 Munich, Germany\\
           \email{alin.albu-schaeffer@dlr.de}
}

\def\makeheadbox{
	\hbox to0pt{\vbox{\baselineskip=10dd\hrule\hbox to\hsize{\vrule\kern3pt\vbox{\kern3pt
    \hbox{\bfseries International Journal of Computer Vision}
	\hbox{This is a pre-peer review, pre-print version of the article}
	\kern3pt}\hfil\kern3pt\vrule}\hrule}%
	\hss}
}

\date{}

\graphicspath{{illustrations/}}

\begin{acronym}
	\acro{ICP}[ICP]{Iterative Closest Point}
	\acro{PDF}[PDF]{probability density function}
	\acro{AUC}[AUC]{area under curve}
	\acro{OPT}[OPT]{Object Pose Tracking}
	\acro{RBOT}[RBOT]{Region-Based Object Tracking}
	\acro{6DoF}[6DoF]{six degrees of freedom}
	\acro{SLAM}[SLAM]{Simultaneous Localization and Mapping}
	\acro{CNN}[CNN]{convolutional neural network}
\end{acronym}

\maketitle

\begin{abstract}
	Region-based methods have become increasingly popular for model-based, monocular 3D tracking of texture-less objects in cluttered scenes.
	However, while they achieve state-of-the-art results, most methods are computationally expensive, requiring significant resources to run in real-time.
	In the following, we build on our previous work and develop \textit{SRT3D}, a sparse region-based approach to 3D object tracking that bridges this gap in efficiency.
	Our method considers image information sparsely along so-called correspondence lines that model the probability of the object's contour location.
	We thereby improve on the current state of the art and introduce smoothed step functions that consider a defined global and local uncertainty.
	For the resulting probabilistic formulation, a thorough analysis is provided.
	Finally, we use a pre-rendered sparse viewpoint model to create a joint posterior probability for the object pose.
	The function is maximized using second-order Newton optimization with Tikhonov regularization.
	During the pose estimation, we differentiate between global and local optimization, using a novel approximation for the first-order derivative employed in the Newton method.
	In multiple experiments, we demonstrate that the resulting algorithm improves the current state of the art both in terms of runtime and quality, performing particularly well for noisy and cluttered images encountered in the real world.
	
\end{abstract}

\section{Introduction}\label{sec:r}
Tracking a rigid object in 3D space and predicting its \ac{6DoF} pose is an essential task in computer vision.
Its application ranges from augmented reality, where the location of objects is needed to superimpose digital information, to robotics, where the object pose is required for robust manipulation in unstructured environments.
Given consecutive image frames, the goal of 3D object tracking is to estimate both the rotation and translation of a known object relative to the camera.
In contrast to object detection, tracking continuously provides information, which, for example, allows robots to react to unpredicted changes in the environment using visual servoing.
While the problem has been thoroughly studied, many challenges such as partial occlusions, appearance changes, motion blur, background clutter, and real-time requirements still exist.
In this section, we first provide an overview of common techniques.
This is followed by a survey of related work on region-based methods.
Finally, we introduce our approach and summarize the contributions to the current state of the art.

\subsection{3D Object Tracking}\label{ssec:r0}
In the past, many approaches to 3D object tracking have been proposed.
Based on surveys \citep{Lepetit2005, Yilmaz2006}, as well as on recent developments, techniques can be differentiated by their use of key-points, explicit edges, direct optimization, deep learning, depth information, and image regions.
Key-point features such as \textit{SIFT} \citep{Lowe2004}, \textit{ORB} \citep{Rublee2011}, or \textit{BRISK} \citep{Leutenegger2011} have been widely used for 3D object tracking \citep{Wagner2010,Vacchetti2004}, with more recent developments like \textit{LIFT} \citep{Yi2016} and \textit{SuperGlue} \citep{Sarlin2020} introducing deep learning at various stages.
Explicit edges provide an additional source of information that is used by many approaches \citep{Huang2020, Bugaev2018, Seo2014, Comport2006, Drummond2002b, Harris1990}.
Also, direct methods \citep{Engel2018,Seo2016,Crivellaro2014}, which optimize a photometric error and can be traced back to \citet{Lucas1981}, have been proposed.
While all three classes of techniques have valid applications, unfortunately, they also have significant drawbacks.
First, approaches based on key-points and direct optimization require rich texture, limiting the range of suitable objects.
At the same time, edge-based methods, which perform better for low-textured objects, often fail in cluttered scenes.
Finally, motion blur changes the appearance of both texture and edges, leading to additional problems.

To overcome these issues and train the algorithm on data, recently, deep-learning-based approaches that use \acp{CNN} to consider full image information have been proposed.
While they achieve good results, only few approaches \citep{Wen2020} run in real-time, with most methods \citep{Deng2021, Wang2019b, Li2018, Xiang2018, Garon2017} reporting less than 30 frames per second.
However, even the most efficient algorithms require significant resources from high-end GPUs.
In addition, typical disadvantages include time-consuming training and the requirement for a textured 3D model.
Another relatively new development is the availability of affordable depth cameras that measure the surface distance for each pixel.
While purely depth-based object tracking is possible, most methods \citep{Ren2017, Kehl2017, Tan2017, Krull2015, Krainin2011} combine information from both depth and RGB cameras.
In general, this leads to superior results.
Unfortunately, in many applications, using an additional depth sensor is not an option.
Also, note that algorithms require images with high quality.
Depending on hardware, surface distances, surface characteristics, and lighting conditions, such images are hard to obtain.

Because of the discussed shortcomings, region-based techniques \citep{Stoiber2020b, Zhong2020, Tjaden2018, Prisacariu2012} have become increasingly popular.
The big advantage of such methods is that they are able to reliably track a wide variety of objects in cluttered scenes, using only a monocular RGB camera and a texture-less 3D model of the object.
The main assumption is thereby that objects are distinguishable from the background.
As a consequence, no object texture is needed.
While past approaches were computationally expensive, our sparse formulation overcomes this disadvantage.
Finally, based on our experience, region-based methods are robust to motion blur, making it possible to track fast-moving objects.
Because of these excellent properties, the following work focuses on region-based techniques.

\begin{figure*}[t]
	\centering
	\input{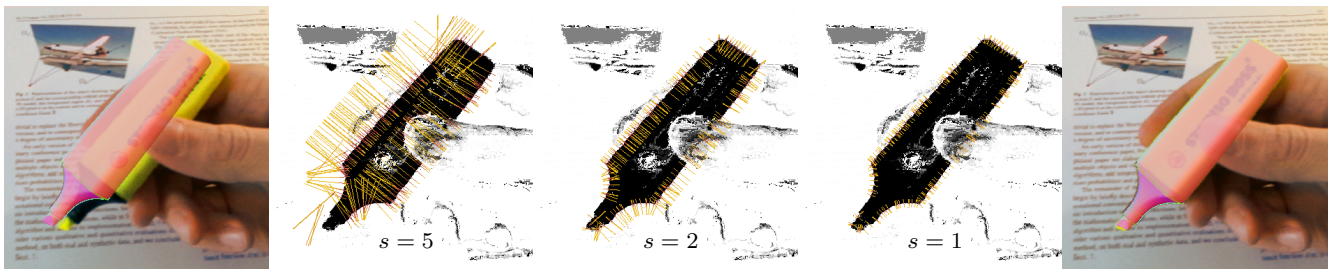}
	\caption{
		Tracking of a marker pen in the real world.
		The image on the left shows a rendered overlay of the object model for the initial pose.
		The estimated pose after the optimization is visualized in the image on the right.
		The three illustrations in the middle show yellow correspondence lines for different scales $s$.
		High probabilities for the contour location are illustrated in red.
		Pixel-wise posterior probabilities that describe the probability that a pixel belongs to the background are encoded in grayscale images.
		Note that during tracking, pixel-wise posteriors are only calculated along correspondence lines.
	}\label{fig:r20}
\end{figure*}

\subsection{Related Work}\label{ssec:r1}
Region-based methods use image statistics to differentiate between a foreground region that corresponds to the object and a background region.
Typically, color statistics are used to model the membership of each pixel.
Based on the two regions, the goal is to find the object pose and corresponding silhouette that best explains the segmentation of the image.
The great potential of this technique was already demonstrated by early approaches that allowed robust tracking in many challenging scenarios \citep{Schmaltz2012, Brox2010, Rosenhahn2007}.
Segmentation and pose tracking were thereby treated as independent problems, with an initial step to extract the contour and a subsequent optimization to find the pose.
\citet{Dambreville2008} later combined the two processes in a single energy function, leading to improved tracking robustness.
Building on this approach and including the pixel-wise posterior membership of \citet{Bibby2008}, \citet{Prisacariu2012} developed \textit{PWP3D}, a real-time-capable algorithm that uses a level-set pose embedding.
It is the foundation of almost all state-of-the-art region-based methods.

Based on \textit{PWP3D}, multiple algorithms were suggested that incorporate additional information, extend the segmentation model, or improve efficiency.
For the combination of both depth- and region-based information, \citet{Kehl2017} extended the energy function of \textit{PWP3D} with a term that is based on the \textit{\ac{ICP}} algorithm.
In a different approach, \citet{Ren2017} tightly coupled region and depth information in a probabilistic formulation that uses 3D signed distance functions.
Recently, object texture was considered using direct optimization of pixel intensity values \citep{Liu2020,Zhong2019} or descriptor fields \citep{Liu2021}.
Also, a combination with an edge-based technique that uses a contour-part model was introduced by \citet{Sun2021}.
Later, \citet{Li2021} developed adaptively weighted local bundles that combine region and edge information.
To improve occlusion handling, \citet{Zhong2020a} suggested the use of learning-based object segmentation.
Finally, the incorporation of measurements from a mobile phone's inertial sensor was suggested by \citet{Prisacariu2015}.

To improve segmentation, \citet{Zhao2014} extended the appearance model of \textit{PWP3D} with a boundary term that considers spatial distribution regularities of pixels.
Later, \citet{Hexner2016} proposed the use of local appearance models that were inspired by the localized contours of \citet{Lankton2008}.
The idea was further improved by \citet{Tjaden2018} with the development of temporally consistent local color histograms.
Finally, \citet{Zhong2020} proposed a method that introduces polar-based region partitioning and edge-based occlusion detection.

For better efficiency, \citet{Zhao2014} suggested a particle-filter-like stochastic optimization that initializes a subsequent damped Newton method.
Later, a hierarchical rendering approach that uses the Levenberg-Marquardt algorithm was developed by \citet{Prisacariu2015}.
Also, \citet{Tjaden2018} proposed the use of a Gauss-Newton method to improve convergence.
In addition to optimization, another idea towards better efficiency is the use of simplified signed distance functions \citep{Liu2020}.
A different approach by \citet{Kehl2017} suggested the use of precomputed contour points to represent the object’s 3D geometry and calculate the energy function sparsely along rays.
Finally, in our previous work \citep{Stoiber2020b}, we improved on this idea and developed a sparse approach that is based on correspondence lines, making our algorithm significantly more efficient than the previous state of the art while achieving better tracking results.

\subsection{Contribution}\label{ssec:r2}
Starting from the ideas presented in the previous section, we focus on the development of \textit{SRT3D}, a highly efficient, sparse approach to region-based tracking.
To keep complexity at a minimum, we only use region information and, like \textit{PWP3D}, adopt a global segmentation model.
For our formulation, we build on our previous method and consider image information sparsely along correspondence lines.
Also, Newton optimization with Tikhonov regularization is used to estimate the object pose.
An illustration of the tracking process with converging correspondence lines at different scales is given in Fig.~\ref{fig:r20}.
While the formulation is similar to our previous method \citep{Stoiber2020b}, our main motivation is to advance the approach and the current state of the art using improved uncertainty modeling and better optimization techniques.
In addition, we provide a more formal derivation and analysis of the highly efficient correspondence line model.
In detail, the main contributions of this work are as follows:
\begin{itemize}
	\item A formal definition of correspondence lines and a thorough mathematical derivation of the probabilistic model that describes the contour location.
	\item Novel smoothed step functions that allow the modeling of both local and global uncertainty.
	\item A detailed theoretical analysis that shows how different parameter settings affect the characteristics of posterior probability distributions.
	\item Global and local optimization strategies and a new approximation for the local first-order derivative.
\end{itemize}

In the remainder, we first provide a detailed derivation of the correspondence line model.
This is followed by the development of a 3D tracking approach that combines the correspondence line model with a sparse representation of the 3D object geometry.
Subsequently, implementation details for the resulting algorithm are discussed.
Finally, we conduct a thorough evaluation on the \textit{\acs{RBOT}} and the \textit{\acs{OPT}} dataset, showing that our approach outperforms the current state of the art by a considerable margin in terms of efficiency and quality.

\section{Correspondence Line Model}\label{sec:p}
In this section, we first provide a formal mathematical definition of correspondence lines.
This is followed by a probabilistic model that considers the segmentation of a correspondence line into foreground and background.
To improve computational efficiency, we extend this model and provide a discrete scale-space formulation.
Finally, we introduce novel smoothed step functions and discuss how their configuration affects the contour location's posterior probability.

\subsection{Correspondence Lines}\label{ssec:p2}
\begin{figure}[t]
	\centering
	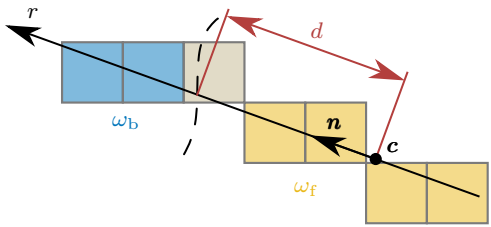
	\caption{
		Correspondence line defined by a center $\pmb{c}$ and a normal vector $\pmb{n}$.
		The illustration shows pixels along the correspondence line as well as the foreground region $\omega_\textrm{f}$ in yellow and the background region $\omega_\textrm{b}$ in blue.
		The contour distance $d$ points from the correspondence line center to an estimated contour, indicated by a dashed line.
	}\label{fig:p20}
\end{figure}

In contrast to most state-of-the-art algorithms, we do not consider image information densely over the entire image.
Instead, inspired by \textit{RAPID} \citep{Harris1990}, pixel values are processed sparsely along correspondence lines \citep{Stoiber2020b}.
The name correspondence line is motivated by the term \textit{correspondence point} used in \textit{\ac{ICP}} \citep{Besl1992}.
Similar to \textit{\ac{ICP}}, correspondences are first defined and the optimization with respect to them is then conducted in a second step.
While for \textit{\ac{ICP}}, individual 3D points are used as data, multiple pixel values along a line are considered in this case.
A visualization of a single correspondence line is shown in Fig.~\ref{fig:p20}.

Starting from our earlier work \citep{Stoiber2020b} and inspired by the commonly used definition of images as $\pmb{I} \colon \pmb{\Omega} \to \{0, \dots, 255\}^3$, we formally denote a correspondence line as a map $\pmb{l}:\omega \to \{0, \dots, 255\}^3$.
In this notation $\pmb{\Omega} \subset \mathbb{R}^2$ describes the image domain while $\omega \subset \mathbb{R}$ is considered the correspondence line domain.
Image values $\pmb{y}$, which are typically accessed using the image coordinate $\pmb{x} = \begin{bmatrix} x& y\end{bmatrix}^\top$ and the image function $\pmb{y} = \pmb{I}(\pmb{x})$, are described using the line coordinate $r$ and the correspondence line function $\pmb{y} = \pmb{l}(r)$.
Correspondence lines are located in the image and remain fixed once they have been established.
The location and orientation of each correspondence line is defined by a center $\pmb{c} = \begin{bmatrix} c_{x} & c_{y}\end{bmatrix}^\top \in \mathbb{R}^2$ in image coordinates and a normal vector $\pmb{n} = \begin{bmatrix} n_{x} & n_{y}\end{bmatrix}^\top \in \mathbb{R}^2$, with $\lVert\pmb{n}\rVert_2 = 1$.
Using this definition, the relation between an image $\pmb{I}$ and a correspondence line $\pmb{l}$ is expressed as follows
\begin{equation} \label{eq:p20}
	\pmb{l}(r) = \pmb{I}(\pmb{c} + r \pmb{n}),
\end{equation}
where image coordinates in $\pmb{I}$ are rounded to the center of the next closest pixel.

\subsection{Probabilistic Model}\label{ssec:p3}
Inspired by the generative model of \citet{Bibby2008}, we derive a probabilistic model for the segmentation of a correspondence line into a foreground region $\omega_\textrm{f}$ and a background region $\omega_\textrm{b}$.
Note that this is the 1D equivalent of the segmentation of a 2D image into the regions $\Omega_\textrm{f}$ and $\Omega_\textrm{b}$.
We assume that there is only one transition between foreground and background.
The location of this transition relative to the line center $\pmb{c}$ is described by the contour distance $d \in \mathbb{R}$.
A visualization of the contour distance is shown in Fig.~\ref{fig:p20}.

To derive the probabilistic model, we first describe the formation process for a single pixel on the correspondence line.
The joint probability distribution writes thereby as follows
\begin{equation} \label{eq:p30}
	p(r, \pmb{y}, d, m) = p(r\mid d, m) p(\pmb{y} \mid  m) p(m) p(d),
\end{equation}
where $m \in \{m_\textrm{f}, m_\textrm{b}\}$ is the model parameter that can denote either foreground or background.
If we condition this distribution on the image value $\pmb{y}$, we obtain
\begin{equation} \label{eq:p31}
	p(r, d, m \mid  \pmb{y}) = p(r\mid d, m) p(m \mid \pmb{y}) p(d).
\end{equation}

Following \citet{Bibby2008}, we use Bayes' theorem and the marginalization over $m$ to calculate the pixel-wise posterior probability
\begin{equation} \label{eq:p33}
	p(m_i \mid  \pmb{y}) = \frac{p(\pmb{y}\mid m_i) p(m_i)}{\sum_{j\in\{\textrm{f},\textrm{b}\}}p(\pmb{y} \mid  m_j)p(m_j)} , \quad i\in\{\textrm{f}, \textrm{b}\},
\end{equation}
where $p(\pmb{y} \mid  m_\textrm{f})$ and $p(\pmb{y} \mid  m_\textrm{b})$ are probability distributions that describe how likely it is that a specific color value is part of the foreground region or the background region, respectively.
The two distributions can be estimated by calculating two color histograms, one over the foreground region and one over the background region.
A detailed explanation of their computation is given in Sect.~\ref{ssec:i1}.
Using the knowledge that foreground and background are equally likely along the correspondence line, i.e. $ p(m_\textrm{f}) =p(m_\textrm{b})$, we obtain
\begin{equation} \label{eq:p34}
	p(m_i \mid  \pmb{y}) = \frac{p(\pmb{y}\mid m_i)}{p(\pmb{y} \mid  m_\textrm{f}) + p(\pmb{y} \mid  m_\textrm{b})} , \quad i\in\{\textrm{f}, \textrm{b}\}.
\end{equation}

Finally, based on Eq.~\eqref{eq:p31}, we are able to marginalize over $m$ and condition on $r$ to express the posterior probability for the contour distance $d$ as
\begin{equation} \label{eq:p32}
	p(d\mid  r, \pmb{y}) = \frac{1}{p(r)} \sum_{i \in \{\textrm{f}, \textrm{b}\}}p(r \mid  d, m_i) p(m_i \mid  \pmb{y}) p(d).
\end{equation}

To calculate the posterior probability over the entire correspondence line domain $\omega$, we assume pixel-wise independence and, based on Eq.~(\ref{eq:p32}), write
\begin{equation} \label{eq:p35}
	p(d\mid  \omega, \pmb{l}) \propto \prod_{r\in\omega}\sum_{i \in \{\textrm{f}, \textrm{b}\}}p(r \mid  d, m_i) p(m_i \mid  \pmb{l}(r)).
\end{equation}
Note that $p(r)$ and $p(d)$ are considered to be uniform and constant and are thus dropped.
Also, while pixel-wise independence does not hold in general, it is a well-established approximation that allows us to avoid ill-defined assumptions for spatial regularities and is close enough to reality to yield good results.
The conditional line coordinate probability $p(r\mid d,m)$ will be discussed in Sect.~\ref{ssec:p5}.
Similar to the probabilistic model of \citet{Bibby2008}, which describes the probability of a shape kernel given information from an image, Eq.~(\ref{eq:p35}) provides the probability of the contour distance $d$ given data from a correspondence line.

\subsection{Discrete Scale-Space Formulation}\label{ssec:p4}
Estimating the distribution of posterior probabilities is computationally expensive since, for each distance $d$, the product in Eq.~(\ref{eq:p35}) has to be computed over the entire domain $\omega$.
This results in quadratic complexity for the calculation of the entire distribution.
In contrast, pixel-wise posterior probabilities $p(m\mid \pmb{y})$ are used in the posterior probability calculation of multiple distances $d$, leading to linear complexity.
Shifting computation from the calculation of the distribution to the calculation of pixel-wise posterior probabilities thus allows us to improve computational efficiency.
Also, it is advantageous to normalize correspondence lines in a way that ensures that a line coordinate pointing to a segment center for one correspondence line points to a segment center for all correspondence lines.
This uniformity can be used in the precalculation of smoothed step function values to further improve efficiency.

\begin{figure}[t]
	\centering
	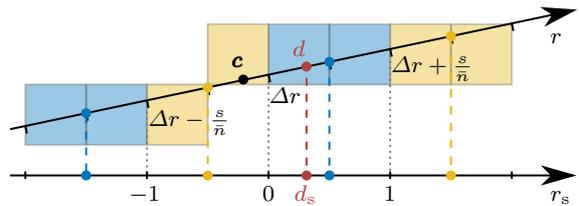
	\caption{
		Example of the relation between the unscaled space $r$ along the correspondence line and the scale-space $r_\textrm{s}$.
		Neighboring pixels that are combined into segments are visualized by the same color in blue or yellow.
		Blue and yellow dots indicate the center of each segment and the corresponding discretized value in the scale-space.
		An example of the contour distance is illustrated in red.
		The offset $\Delta r$ is chosen in a way that ensures that discretized values in the scale-space are the same for all correspondence lines.
		In this example, $\Delta r$ points to the closest edge between pixels.
	}\label{fig:p40}
\end{figure}

In the following, we thus adopt the discrete scale-space formulation from our previous method \citep{Stoiber2020b} to combine multiple pixels into segments.
In addition, the formulation projects from the continuous space along the correspondence line into a discrete space that is independent of a correspondence line's location and orientation.
An illustration of this transformation is shown in Fig.~\ref{fig:p40}.
Both line coordinates and contour distances are projected as follows
\begin{gather}\label{eq:p40}
	r_\textrm{s} = (r - \Delta r) \frac{\bar{n}}{s},\\\label{eq:p40_0}
	d_{\textrm{s}} = (d - \Delta r) \frac{\bar{n}}{s},
\end{gather}
with $s \in \mathbb{N}^+$ the scale that describes the number of pixels combined into a segment, $\bar{n} = \max(|n_{x}|,|n_{y}|)$ the major absolute normal component that projects a correspondence line to the closest horizontal or vertical image coordinate, and $\Delta r \in \mathbb{R}$ the offset from the correspondence line center $\pmb{c}$ to a defined pixel location.

Based on Eq.~(\ref{eq:p35}), the posterior probability in the discrete scale-space is calculated as
\begin{equation} \label{eq:p41}
	p(d_\textrm{s}\mid  \omega_\textrm{s}, \pmb{l}_\textrm{s}) \propto \prod_{r_\textrm{s} \in \omega_\textrm{s}}\sum_{i \in \{\textrm{f}, \textrm{b}\}}p(r_\textrm{s} \mid  d_\textrm{s}, m_i) p(m_i \mid  \pmb{l}_\textrm{s}(r_\textrm{s})),
\end{equation}
where $\omega_\textrm{s}$ is the scaled correspondence line domain and $\pmb{s} = \pmb{l}_\textrm{s}(r_\textrm{s})$ a set-valued function that maps from the scaled line coordinate $r_\textrm{s}$ to the segment $\pmb{s}$, which is a set of the closest $s$ pixel values $\pmb{y}$.
Similar to pixel-wise posteriors in Eq.~(\ref{eq:p34}) and assuming pixel-wise independence, segment-wise posteriors are defined as
\begin{equation} \label{eq:p42}
	p(m_i \mid  \pmb{s}) = \frac{\prod\limits_{\pmb{y} \in \pmb{s}} p(\pmb{y}\mid m_i)} {\prod\limits_{\pmb{y} \in  \pmb{s}} p(\pmb{y} \mid  m_\textrm{f}) + \prod\limits_{\pmb{y} \in  \pmb{s}} p(\pmb{y} \mid  m_\textrm{b})} , \quad i\in\{\textrm{f}, \textrm{b}\}.
\end{equation}
The derived formulation allows to efficiently cover the correspondence line domain $\omega$, using the scale parameter $s$ to set the segment size and to adjust between accuracy and efficiency.
In the following, we will again drop the index $\textrm{s}$ for all variables to simplify the notation.
Note, however, that all definitions and derivations are valid both for the original space and for the discrete scale-space formulation.

\subsection{Smoothed Step Functions}\label{ssec:p5}
\begin{figure}[t]
	\centering
	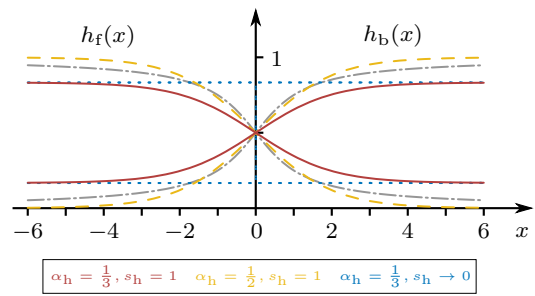
	\caption{
		Smoothed step functions $h_\textrm{f}$ and $h_\textrm{b}$ that model the conditional line coordinate probabilities $p(r\mid d,m_\textrm{f})$ and $p(r\mid d,m_\textrm{b})$.
		The functions $h_\textrm{f}(x) = \frac{1}{2} - \frac{1}{\pi}\tan^{-1}\big(\frac{x}{s_\textrm{h}}\big)$ and $h_\textrm{b}(x) = \frac{1}{2} + \frac{1}{\pi}\tan^{-1}\big(\frac{x}{s_\textrm{h}}\big)$ used by \citet{Zhong2020} and \citet{Tjaden2018} are illustrated by dash-dotted gray lines.
		The definitions $h_\textrm{f}(x) = \frac{1}{2} - \frac{1}{2}\tanh\big(\frac{x}{2s_\textrm{h}}\big)$ and $h_\textrm{b}(x) = \frac{1}{2} + \frac{1}{2}\tanh\big(\frac{x}{2s_\textrm{h}}\big)$ from our previous work \citep{Stoiber2020b} are shown as dashed yellow lines.
		In both plots, the slope parameter $s_\textrm{h}=1$ is used.
		For the proposed functions in Eq.~(\ref{eq:p50}) and (\ref{eq:p51}), solid red lines correspond to $\alpha_\textrm{h} = \frac{1}{3}$ and $s_\textrm{h} = 1$, while dotted blue lines show the functions for $\alpha_\textrm{h} = \frac{1}{3}$ and $s_\textrm{h} \to 0$.
		In addition to visualizing the definitions from our previous work, the dashed yellow lines illustrate the proposed functions for $\alpha_\textrm{h} = \frac{1}{2}$ and $s_\textrm{h}=1$.
	}\label{fig:p50}
\end{figure}

To model the conditional probabilities of the line coordinate $p(r\mid d,m_\textrm{f})$ and $p(r\mid d,m_\textrm{b})$, different smoothed step functions $h_\textrm{f}$ and $h_\textrm{b}$ have been used.
While most state-of-the-art algorithms \citep{Zhong2020, Tjaden2018} use a function based on the arctangent, we previously proved that a hyperbolic tangent results in a Gaussian distribution for the posterior probability $p(d\mid \omega, \pmb{l})$ \citep{Stoiber2020b}.
In both functions, the smoothed slope is used to model a local uncertainty with respect to the exact location of the foreground and background transition.
Considering the plots of the two models in Fig.~\ref{fig:p50}, one notices that the functions quickly converge towards either zero or one for increasing absolute values of $x = r - d$.
Except for a small area around zero, both models thus assume that, given the model $m$ and the contour distance $d$, one knows perfectly on which side of the contour the line coordinate $r$ lies.
In the following, we will argue that for real-world applications, this assumption is wrong.

While the pixel-wise posterior probability in Eq.~(\ref{eq:p34}) provides very good predictions for the model $m$, it is still an imperfect simplification of the real world.
Typical effects that are not considered by the statistical model are image noise or fast appearance changes that can lead to pixel colors that are not yet present in the color histograms.
Another effect originates from pixels that are wrongly classified due to imperfect segmentation and that are then assigned to the wrong color histograms.
Finally, there also remains the question if a statistical model that purely relies on pixel colors is sufficient to capture all the statistical effects in the real world and is able to perfectly predict the model $m$.

To take those limitations into account and consider a constant, global uncertainty in the foreground and background model, we extend the formulation from our previous work \citep{Stoiber2020b} and propose the following functions
\begin{gather}\label{eq:p50}
	h_\textrm{f}(x) = \frac{1}{2} - \alpha_\textrm{h}\tanh\bigg(\frac{x}{2s_\textrm{h}}\bigg),\\\label{eq:p51}
	h_\textrm{b}(x) = \frac{1}{2} + \alpha_\textrm{h}\tanh\bigg(\frac{x}{2s_\textrm{h}}\bigg).
\end{gather}
Note that the amplitude parameter $\alpha_\textrm{h} \in [0,0.5]$ was added to the original definitions that only considered the slope parameter $s_\textrm{h} \in \mathbb{R}^+$.
For $\alpha_\textrm{h} = \frac{1}{2}$ the equations are equal to our previous formulation.
Examples of the proposed functions are shown in Fig.~\ref{fig:p50}.

In addition to viewing $\alpha_\textrm{h}$ as a simple amplitude parameter, we are able to demonstrate that there is also another interpretation.
For this, we assume that the model $m$ is extended with a third class $m_\textrm{n}$ that considers external effects that are independent of the foreground and background model $m_\textrm{f}$ and $m_\textrm{b}$.
For this scenario, we can show that $p(m_\textrm{f}) = p(m_\textrm{b}) = \alpha_\textrm{h}$ and that $p(m_\textrm{n}) = 1 - 2\alpha_\textrm{h}$.
Following this interpretation, the amplitude parameter thus allows us to set the probability that a pixel's color is generated by the foreground or background model in contrast to some other effect that is considered as noise.
This again shows that the amplitude parameter $\alpha_\textrm{h}$ is able to model a constant, global uncertainty.
Note that in this scenario, the original smoothed step functions that converge to zero or one are used for $p(r\mid d,m_\textrm{f})$ and $p(r\mid d,m_\textrm{b})$ and a constant function $p(r\mid d,m_\textrm{n}) = \frac{1}{2}$ is adopted for the noise model.
A detailed derivation of this extended model and a proof of its equivalence to the use of the functions in Eq.~(\ref{eq:p50}) and (\ref{eq:p51}) is given in Appendix~\ref{sec:aa}.

\subsection{Posterior Probability Distribution}\label{ssec:p6}
Given the smoothed step functions $h_\textrm{f}$ and $h_\textrm{b}$ that model the conditional line coordinate probabilities $p(r\mid d,m_\textrm{f})$ and $p(r\mid d,m_\textrm{b})$, the final expression of the posterior probability distribution from Eq.~(\ref{eq:p35}) can be written as
\begin{equation}\label{eq:p60}
	p(d\mid \omega,\pmb{l}) \propto \prod_{r\in\omega}h_\textrm{f}(r-d)p_\textrm{f}(r) + h_\textrm{b}(r-d)p_\textrm{b}(r),
\end{equation}
with the abbreviations $p_\textrm{f}(r) = p(m_\textrm{f}\mid \pmb{l}(r))$ and $p_\textrm{b}(r) = p(m_\textrm{b}\mid \pmb{l}(r))$.
In the following, we provide a detailed analysis to understand how the slope parameter $s_\textrm{h}$ and the amplitude parameter $\alpha_\textrm{h}$ affect this distribution.
We thereby assume a contour at the correspondence line center and step functions for the pixel-wise posteriors $p_\textrm{f}$ and $p_\textrm{b}$.
Note that the assumption of step functions corresponds well with real-world experiments that show that, in most cases, there is a distinct split between foreground and background.

\begin{figure}[t]
	\centering
	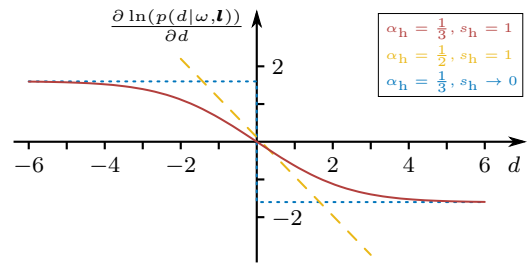
	\caption{
	First-order derivatives of the log-posterior with respect to the contour distance $d$ for different slope and amplitude parameters $s_\textrm{h}$ and $\alpha_\textrm{h}$.
	The solid red line shows the derivative for $\alpha_\textrm{h} = \frac{1}{3}$ and $s_\textrm{h} = 1$, which yields a function with a smooth transition from an upper bound to a lower bound.
	The dashed yellow line shows the function for $\alpha_\textrm{h} = \frac{1}{2}$ and $s_\textrm{h} = 1$.
	This produces a linear first-order derivative.
	Finally, using $\alpha_\textrm{h} = \frac{1}{3}$ and $s_\textrm{h} \to 0$ results in a perfect step function illustrated by the dotted blue line.
	}\label{fig:p60}
\end{figure}

For the analysis, we start with the calculation of the first-order derivative of the log-posterior with respect to the contour distance $d$.
The derivation is conducted similar to our previous work \citep{Stoiber2020b} and assumes continuous functions with infinitesimally small pixels.
Based on a detailed derivation given in Appendix~\ref{sec:ab}, the closed-form solution is written as
\begin{equation}\label{eq:p61}
	\frac{\partial\ln\big(p(d\mid \omega,\pmb{l})\big)}{\partial d} = -2\tanh^{-1}\bigg(2 \alpha_\textrm{h} \tanh\bigg(\frac{d}{2s_\textrm{h}}\bigg)\bigg).
\end{equation}
A visualization of this function for different slope and amplitude parameters $\alpha_\textrm{h}$ and $s_\textrm{h}$ is given in Fig.~\ref{fig:p60}.
The plot shows that the amplitude parameter $\alpha_\textrm{h}$ controls not only the amplitude of $h_\textrm{f}$ and $h_\textrm{b}$ but also the amplitude of the first-order derivative.
For $\alpha_\textrm{h} = \frac{1}{2}$, the first-order derivative converges to a linear function.
At the same time, the parameter $s_\textrm{h}$ affects both the slope of $h_\textrm{f}$ and $h_\textrm{b}$ and the slope of the first-order derivative.
For $s_\textrm{h} \to 0$ it leads to a perfect step function.

\begin{figure}[t]
	\centering
	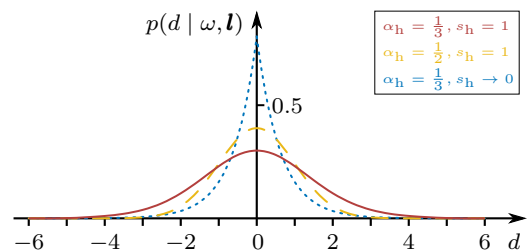
	\caption{
		Posterior probability distributions for different slope and amplitude parameters $s_\textrm{h}$ and $\alpha_\textrm{h}$.
		The solid red line shows the function for $\alpha_\textrm{h} = \frac{1}{3}$ and $s_\textrm{h} = 1$, which leads to a very flat distribution.
		Note that the function was computed numerically.
		Using $\alpha_\textrm{h} = \frac{1}{2}$ and $s_\textrm{h} = 1$ results in a Gaussian distribution shown by the dashed yellow line.
		The parameters $\alpha_\textrm{h} = \frac{1}{3}$ and $s_\textrm{h} \to 0$ yield a Laplace distribution for the posterior probability that is illustrated by a dotted blue line.
	}\label{fig:p61}
\end{figure}

For the two edge cases with $\alpha_\textrm{h} = \frac{1}{2}$ and $s_\textrm{h} \to 0$, Eq.~(\ref{eq:p61}) can be simplified, and we are able to calculate a closed-form solution for the posterior probability distribution.
In the case of $\alpha_\textrm{h} = \frac{1}{2}$, for which we obtain the smoothed step functions of our previous approach \citep{Stoiber2020b}, the posterior probability distribution results in a perfect Gaussian
\begin{equation}\label{eq:p62}
p(d\mid \omega,\pmb{l}) = \frac{1}{\sqrt{2\pi s_\textrm{h}}}\exp\bigg(-\frac{d^2}{2s_\textrm{h}}\bigg),
\end{equation}
where the slope parameter $s_\textrm{h}$ is equal to the variance.
In the case of $s_\textrm{h} \to 0$, which leads to sharp step functions for $h_\textrm{f}$ and $h_\textrm{b}$, the posterior probability distribution becomes a perfect Laplace distribution
\begin{equation}\label{eq:p63}
	p(d\mid \omega,\pmb{l}) = \frac{1}{2b}\exp\bigg(-\frac{|d|}{b}\bigg),\quad b=\frac{1}{2\tanh^{-1}(2\alpha_\textrm{h})},
\end{equation}
where $b \in \mathbb{R}^+$ is the scale parameter of the Laplace distribution that depends on $\alpha_\textrm{h}$.
A detailed derivation of the two functions is provided in Appendix~\ref{sec:ac}.
Examples for both distributions, as well as a mixed posterior distribution with $s_\textrm{h}=1$ and $\alpha_\textrm{h}=\frac{1}{3}$, are visualized in Fig.~\ref{fig:p61}.
The plot shows that while the Laplace distribution has a pronounced peak, the Gaussian distribution has a smoothed maximum for which nearby values have similarly high probabilities.
This coincides with our intuition that the slope parameter $s_\textrm{h}$ controls local uncertainty, allowing multiple values $d$ to be almost equally likely.
At the same time, the amplitude parameter $\alpha_\textrm{h}$ controls the size of the peak compared to its surroundings, thereby controlling global uncertainty.
Combining the two parameters in a mixed distribution results in a function that is able to consider both local and global uncertainty simultaneously.
Given the detailed knowledge about correspondence lines and the posterior probability distribution, we are now able to develop \textit{SRT3D}, a highly efficient, sparse approach to region-based 3D object tracking.

\section{Region-Based 3D Tracking}\label{sec:t}
In this section, we first define basic mathematical concepts.
This is followed by the description of a sparse viewpoint model, which avoids the rendering of the 3D model during tracking.
Combining this geometry representation with the correspondence line model developed in the previous section, we are able to formulate a joint posterior probability with respect to the pose.
The probability is maximized using Newton optimization with Tikhonov regularization.
Finally, we define the required gradient vector and Hessian matrix for the Newton method.
We thereby differentiate between global and local optimization to ensure both fast convergence and high accuracy.

\subsection{Preliminaries}\label{ssec:t0}
In the following work, we define 3D model points as $\pmb{X} = \begin{bmatrix} X& Y& Z\end{bmatrix}^\top\in \mathbb{R}^3$ and use the tilde notation to write the homogeneous form $\pmb{\widetilde{X}} = \begin{bmatrix} X& Y& Z& 1\end{bmatrix}^\top$.
For the projection of a 3D model point $\pmb{X}$ into the image space, we assume an undistorted image and use the pinhole camera model
\begin{equation}\label{eq:t01}
	\pmb{x} = \pmb{\pi}(\pmb{X}) =
	\begin{bmatrix}
		\frac{X}{Z}f_x + p_x\\
		\frac{Y}{Z}f_y + p_y
	\end{bmatrix},
\end{equation}
with $f_x$ and $f_y$ the focal lengths and $p_x$ and $p_y$ the principal point coordinates given in units of pixels.
The inverse operation, which is the reconstruction of a 3D model point from an image coordinate $\pmb{x}$ and corresponding depth value $d_Z$ along the optical axis, can be written as
\begin{equation}\label{eq:t01_2}
	\pmb{X} = \pmb{\pi}^{-1}(\pmb{x}, d_Z) = d_Z
	\begin{bmatrix}
		\frac{x-p_x}{f_x}\\
		\frac{y-p_y}{f_y}\\
		1
	\end{bmatrix}.
\end{equation}

\begin{figure}[t]
	\centering
	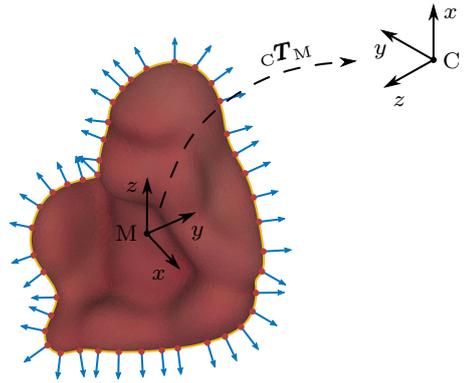
	\caption{
		Illustration of a 2D rendering computed from a 3D mesh model.
		The model reference frame $\textrm{M}$ is shown at the center of the object, while a camera reference frame $\textrm{C}$ is shown at the right upper corner of the image.
		The transformation from the model to the camera reference frame that is described by ${}_\textrm{C}\pmb{T}_\textrm{M}$ is indicated by a dashed arrow.
		The contour of the rendered model is highlighted by a yellow line.
		Red points and blue arrows illustrate 2D contour points and approximated normal vectors.
	}\label{fig:t00}
\end{figure}

To describe the relative pose between the model reference frame $\textrm{M}$ and the camera reference frame $\textrm{C}$, we use the homogeneous matrix ${}_\textrm{C}\pmb{T}_\textrm{M} \in \mathbb{SE}(3)$.
For the transformation of a 3D model point, we can then write
\begin{equation} \label{eq:t03}
	_\textrm{C}\pmb{\widetilde{X}} = {}_\textrm{C}\pmb{T}_\textrm{M}{}_\textrm{M}\pmb{\widetilde{X}} =
	\begin{bmatrix}
		_\textrm{C}\pmb{R}_\textrm{M} & _\textrm{C}\pmb{t}_\textrm{M} \\ \pmb{0} & 1
	\end{bmatrix}
	{}_\textrm{M}\pmb{\widetilde{X}},
\end{equation}
where $_\textrm{C}\pmb{\widetilde{X}}$ and $_\textrm{M}\pmb{\widetilde{X}}$ are 3D model points written in the camera reference frame $\textrm{C}$ and the model reference frame $\textrm{M}$, respectively, and where $_\textrm{C}\pmb{R}_\textrm{M} \in \mathbb{SO}(3)$ and $_\textrm{C}\pmb{t}_\textrm{M} \in \mathbb{R}^3$ are the rotation matrix and the translation vector that define the transformation from $\textrm{M}$ to $\textrm{C}$.
An illustration of the two reference frames and a homogeneous transformation matrix is given in Fig.~\ref{fig:t00}.

For small variations, the angle-axis representation, which is a minimal representation, is used.
With the exponential map, the rotation matrix writes as
\begin{equation} \label{eq:t04}
	\pmb{R} = \exp([\pmb{r}]_\times) = \pmb{I} + [\pmb{r}]_\times + \frac{1}{2!}[\pmb{r}]_\times^2 + \frac{1}{3!}[\pmb{r}]_\times^3 + ...,
\end{equation}
where $[\pmb{r}]_\times$ is the skew-symmetric matrix of $\pmb{r} \in \mathbb{R}^3$.
By neglecting higher-order terms of the series expansion, Eq.~(\ref{eq:t04}) can be linearized.
We are then able to write the linear variation of a 3D model point in the camera reference frame $\textrm{C}$ as
\begin{equation} \label{eq:t05}
	_\textrm{C}\pmb{\widetilde{X}}(\pmb{\theta}) =
	\begin{bmatrix}
		_\textrm{C}\pmb{R}_\textrm{M} & _\textrm{C}\pmb{t}_\textrm{M} \\ \pmb{0} & 1
	\end{bmatrix}
	\begin{bmatrix}
		\pmb{I} + [\pmb{\theta}_\textrm{r}]_\times & \pmb{\theta}_\textrm{t} \\ \pmb{0} & 1
	\end{bmatrix}
	{}_\textrm{M}\pmb{\widetilde{X}},
\end{equation}
with the rotational variation $\pmb{\theta}_\textrm{r} \in \mathbb{R}^3$, the translational variation $\pmb{\theta}_\textrm{t} \in \mathbb{R}^3$, and the full variation vector $\pmb{\theta}^\top = \begin{bmatrix} \pmb{\theta}_\textrm{r}^\top & \pmb{\theta}_\textrm{t}^\top \end{bmatrix}$.
Note that, since the object is typically moved significantly more than the camera, it is more natural to variate 3D points in the model reference frame $\textrm{M}$ instead of the camera reference frame $\textrm{C}$.
Also, the variation in the model reference frame has the advantage that a simple extension of the algorithm to multiple cameras is possible.

\subsection{Sparse Viewpoint Model}\label{ssec:t1}
In contrast to most state-of-the-art region-based methods, we do not use the 3D geometry in the form of a mesh model directly.
Instead, similar to \citet{Tan2017}, we employ a representation that we call a sparse viewpoint model.
To create this model, the 3D geometry is rendered from a number of $n_\textrm{v}$ viewpoints all around the object.
Virtual cameras are thereby placed on the vertices of a geodesic grid that surrounds the object.
For each rendering, $n_\textrm{c}$ points $\pmb{x}_i\in\mathbb{R}^2$ are randomly sampled from the contour of the model.
Subsequently, the vectors $\pmb{n}_i\in\mathbb{R}^2$ that are normal to the contour are approximated for each point.
Note that $\lVert\pmb{n}_i\rVert_2 = 1$.
An illustration of a rendering with sampled 2D contour points and normal vectors is shown in Fig.~\ref{fig:t00}.
Based on the 2D entities, 3D vectors with respect to the model reference frame are then reconstructed as follows
\begin{gather} \label{eq:t10}
	{}_\textrm{M}\pmb{\widetilde{X}}_i = {}_\textrm{M}\pmb{T}_\textrm{C} \ \pmb{\tilde{\pi}}^{-1}(\pmb{x}_i, d_{Zi}),\\
	{}_\textrm{M}\pmb{N}_i = {}_\textrm{M}\pmb{R}_\textrm{C}
	\begin{bmatrix}
		\pmb{n}_i\\
		0
	\end{bmatrix},
\end{gather}
where the tilde notation in $\pmb{\tilde{\pi}}^{-1}$ indicates that the 3D model point is returned in homogeneous form and $d_{Zi}$ is the depth value from the rendering.
Note that in this case, $\textrm{C}$ denotes the reference frame of the virtual camera from which the rendering was created.
In addition to those vectors, we also compute the orientation vector ${}_\textrm{M}\pmb{v} = {}_\textrm{M}\pmb{R}_\textrm{C}\pmb{e}_\textrm{Z}$ that points from the camera to the model center, where $\pmb{e}_\textrm{Z} = \begin{bmatrix}0&0&1\end{bmatrix}^\top$.
The computed 3D model points, normal vectors, and the orientation vector are then stored for each view.

The sparse viewpoint model allows for a highly efficient representation of the model contour.
Given a specific pose with ${}_\textrm{M}\pmb{R}_\textrm{C}$ and ${}_\textrm{C}\pmb{t}_\textrm{M}$, the process of rendering the model and computing the contour reduces to a simple search for the closest precomputed view $i_\textrm{v}$
\begin{equation}\label{eq:t11}
	i_\textrm{v} = \argmax_{i \in \{1, \dots ,n_\textrm{v}\}}( {}_\textrm{M}\pmb{v}_{i}^\top{}_\textrm{M}\pmb{R}_\textrm{C} {}_\textrm{C}\pmb{t}_\textrm{M}),
\end{equation}
and the subsequent projection of the corresponding 3D model points and normal vectors into the image.
Note that this high efficiency is especially important during the optimization of the joint posterior probability, where the pose changes in each iteration.

\subsection{Joint Posterior Probability}\label{ssec:t2}
In the following, we combine the developed sparse viewpoint model with the correspondence line model from Sect.~\ref{sec:p} to define a joint posterior probability with respect to the pose variation.
However, before probabilities can be calculated, the location and orientation of correspondence lines need to be defined.
For this, 3D model points and normal vectors from the closest view of the sparse viewpoint model are projected into the image using the following equations
\begin{gather}\label{eq:t20}
	\pmb{c}_i = \pmb{\pi}\big({}_\textrm{C}\pmb{T}_\textrm{M} {}_\textrm{M}\pmb{\widetilde{X}}_i\big),\\\label{eq:t20_0}
	\pmb{n}_i \propto \big({}_\textrm{C}\pmb{R}_\textrm{M} {}_\textrm{M}\pmb{N}_i\big)_{2\times 1},
\end{gather}
where the normal vector $\pmb{n}_i$ is normalized to $\lVert\pmb{n}_i\rVert_2=1$ and $()_{2\times1}$ denotes the first two elements of a vector.

\begin{figure}[t]
	\centering
	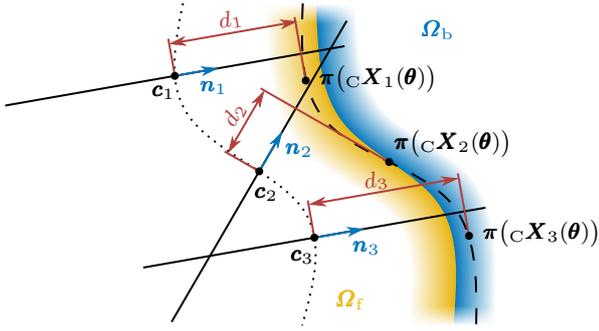
	\caption{
		Correspondence lines defined by a center $\pmb{c}_i$ and a normal vector $\pmb{n}_i$.
		Variated contour distances $d_i$ are measured along the correspondence lines from the centers $\pmb{c}_i$ to the projected 3D model points ${}_\textrm{C}\pmb{X}_i$ that depend all on the same pose variation $\pmb{\theta}$.
		The object contour of the original pose estimate, which was used to define the correspondence lines, is indicated by a dotted line.
		The current estimate of the contour that depends on the pose variation vector $\pmb{\theta}$ is shown by a dashed line.
		The ground truth segmentation that we try to estimate is given by the foreground region $\pmb{\Omega}_\textrm{f}$ in yellow and the background region $\pmb{\Omega}_\textrm{b}$ in blue.
		Note that while contours are illustrated as continuous lines, in our method, they are represented by points and normal vectors from the closest view of the sparse viewpoint model.
	}\label{fig:t20}
\end{figure}

Once all correspondence lines have been defined, we are able to variate the current pose and calculate contour distances $d_i$ with respect to the pose variation vector $\pmb{\theta}$.
Contour distances are thereby calculated as the distances along normal vectors $\pmb{n}_i$ from correspondence line centers $\pmb{c}_i$ to projected 3D model points $\pmb{X}_i$
\begin{equation}\label{eq:t21}
	d_i(\pmb{\theta}) = \pmb{n}_i^\top\big(\pmb{\pi}({}_\textrm{C}\pmb{X}_i(\pmb{\theta}))-\pmb{c}_i\big).
\end{equation}
Note that the same 3D model points $\pmb{X}_i$ are used as for the definition of correspondence lines.
Also, while we do not write this explicitly, 3D model points ${}_\textrm{C}\pmb{X}_i$ and contour distances $d_i$ also depend on the current pose estimate ${}_\textrm{C}\pmb{T}_\textrm{M}$, which might be different from the pose that was used to define correspondence lines.
An example of multiple correspondence lines with variated contour distances is shown in Fig.~\ref{fig:t20}.

Finally, assuming a number of $n_\textrm{c}$ independent correspondence lines and using the discrete scale-space formulation from Sect.~\ref{ssec:p4} to improve efficiency, the joint posterior probability can be calculated as
\begin{equation}\label{eq:t22}
	p(\pmb{\theta}\mid \pmb{\mathcal{D}}) \propto \prod_{i=1}^{n_\textrm{c}}p(d_{\textrm{s}i}(\pmb{\theta})\mid\omega_{\textrm{s}i},\pmb{l}_{\textrm{s}i}),
\end{equation}
where $\pmb{\mathcal{D}}$ describes the data from all correspondence lines.
Note that the transformation of contour distances $d_i$ from the original space to the discrete scale-space is given by Eq.~(\ref{eq:p40_0}).
The developed joint posterior probability describes how well the current pose estimate explains the segmentation of the image into a foreground region, that corresponds to the tracked object, and a background region.

\subsection{Optimization}\label{ssec:t3}
To maximize the joint posterior probability, we estimate the variation vector $\pmb{\hat{\theta}}$ and iteratively update the pose.
For a single iteration, the variation vector is calculated using the Newton method with Tikhonov regularization
\begin{equation} \label{eq:t30}
	\pmb{\hat{\theta}} = \bigg(-\pmb{H} + 
	\begin{bmatrix}
		\lambda_\textrm{r} \pmb{I}_3 & \pmb{0}\\
		\pmb{0} & \lambda_\textrm{t} \pmb{I}_3
	\end{bmatrix}
	\bigg)^{-1}\pmb{g},
\end{equation}
where $\pmb{g}$ is the gradient vector, $\pmb{H}$ is the Hessian matrix, $\pmb{I}_3$ the $3\times 3$ identity matrix, and $\lambda_\textrm{r}$ and $\lambda_\textrm{t}$ are the regularization parameters for rotation and translation, respectively.
The gradient vector and the Hessian matrix are defined as the first- and second-order derivatives of the joint log-posterior
\begin{gather}\label{eq:t31}
	\pmb{g}^\top = \frac{\partial}{\partial\pmb{\theta}} \ln\big(p(\pmb{\theta}\mid\pmb{\mathcal{D}})\big) \Big\vert_{\pmb{\theta}=\pmb{0}},\\
	\pmb{H} = \frac{\partial^2}{\partial\pmb{\theta}^2} \ln\big(p(\pmb{\theta}\mid\pmb{\mathcal{D}})\big)\Big\vert_{\pmb{\theta}=\pmb{0}}.
\end{gather}
Using the logarithm has the advantage that scaling terms vanish and products turn into summations.
Note that the Hessian represents the curvature of the distribution at a specific location, which for Gaussian probability functions is constant and directly corresponds to the negative inverse variance.
Given this probabilistic interpretation, it can be argued that regularization parameters correspond to a prior probability.
This prior controls how much we believe in the previous pose compared to the current estimate described by the gradient and Hessian.
Consequently, for directions in which the Hessian indicates high uncertainty, the regularization helps to keep the optimization stable and to avoid pose changes that are not supported by sufficient data.

Finally, given a robust estimate for the variation vector, the predicted pose can be updated as follows
\begin{equation} \label{eq:t32}
	_\textrm{C}\pmb{T}_\textrm{M} =
	{}_\textrm{C}\pmb{T}_\textrm{M}
	\begin{bmatrix}
		\exp([\pmb{\hat{\theta}}_\textrm{r}]_\times) & \pmb{\hat{\theta}}_\textrm{t} \\ \pmb{0} & 1
	\end{bmatrix}.
\end{equation}
Because of the exponential map, no orthonormalization is necessary.
By iteratively repeating this process, we are able to optimize towards the pose that best explains the segmentation of the image.

\subsection{Gradient and Hessian Approximation}\label{ssec:t4}
In the following, the gradient vector and the Hessian matrix are approximated in a way that ensures both fast convergence and high accuracy.
Using the chain rule, we write
\begin{equation}\label{eq:t40}
	\pmb{g}^\top = \sum_{i=1}^{n_\textrm{c}}
	\frac{\partial\ln\big(p(d_{\textrm{s}i}\mid\omega_{\textrm{s}i},\pmb{l}_{\textrm{s}i})\big)}{\partial d_{\textrm{s}i}}
	\frac{\partial d_{\textrm{s}i}}{\partial {}_\textrm{C}\pmb{X}_{i}}
	\frac{\partial {}_\textrm{C}\pmb{X}_{i}}{\partial \pmb{\theta}}
	\bigg\vert_{\pmb{\theta}=\pmb{0}},
\end{equation}
\begin{equation}\label{eq:t41}
	\begin{split}
		\pmb{H} \approx \sum_{i=1}^{n_\textrm{c}}
		&\frac{\partial^2\ln\big(p(d_{\textrm{s}i}\mid\omega_{\textrm{s}i},\pmb{l}_{\textrm{s}i})\big)}{\partial {d_{\textrm{s}i}}^2}
		\left(
		\frac{\partial d_{\textrm{s}i}}{\partial{}_\textrm{C}\pmb{X}_{i}}\frac{\partial {}_\textrm{C}\pmb{X}_{i}}{\partial \pmb{\theta}}
		\right)^\top\\
		&\left(\frac{\partial d_{\textrm{s}i}}{\partial _\textrm{C}\pmb{X}_{i}}
		\frac{\partial _\textrm{C}\pmb{X}_{i}}{\partial \pmb{\theta}}\right)
		\bigg\vert_{\pmb{\theta}=\pmb{0}}.
	\end{split}
\end{equation}
Note that for the Hessian matrix, second-order partial derivatives with respect to $d_{\textrm{s}i}$ and ${}_\textrm{C}\pmb{X}_{i}$  are neglected.
Resulting errors are left to the iterative nature of the optimization.
Using Eq.~(\ref{eq:t05}), the first-order derivative of the 3D model point ${}_\textrm{C}\pmb{X}_{i}$ is calculated as
\begin{equation}\label{eq:t42}
	\frac{\partial _\textrm{C}\pmb{X}_{i}}{\partial\pmb{\theta}} =
	{}_\textrm{C}\pmb{R}_\textrm{M} 
	\begin{bmatrix}
		-[_\textrm{M}\pmb{X}_i]_\times && \pmb{I}_3
	\end{bmatrix}.
\end{equation}
With respect to the scaled contour distance $d_{\textrm{s}i}$, both Eq.~(\ref{eq:t21}) and (\ref{eq:p40_0}) are used to write
\begin{equation}\label{eq:t43}
	\begin{split}
		\frac{\partial d_{\textrm{s}i}}{\partial _\textrm{C}\pmb{X}_{i}} =
		\frac{\bar{n}_i}{s}
		\frac{1}{{} _\textrm{C}Z_{i}^2}
		\big[
		&\begin{matrix}
			n_{xi} f_x {}_\textrm{C}Z_{i} && n_{yi} f_y {}_\textrm{C}Z_{i}
		\end{matrix}\\
		&\begin{matrix}
			-n_{xi} f_x {}_\textrm{C}X_{i} - n_{yi} f_y {}_\textrm{C}Y_{i}
		\end{matrix}
		\big].
	\end{split}
\end{equation}
For the calculation of the required first- and second-order derivatives of the log-posterior, we differentiate between global and local optimization.
To some extent, this is similar to our previous approach \citep{Stoiber2020b}.
However, in contrast to that work, we propose different approximations for the local optimization.
Also, we either apply global or local optimization and use the same definition of derivatives for all correspondence lines instead of mixing them.

\begin{figure}[t]
	\centering
	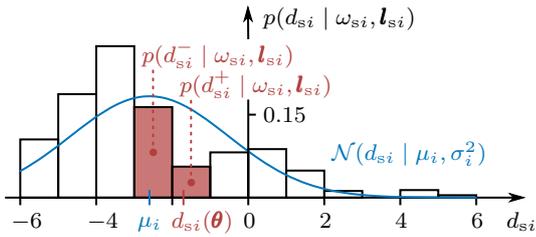
	\caption{
		Discrete posterior probability distribution with noisy measurements.
		For global optimization, the distribution is approximated by a normal distribution $\mathcal{N}(d_{\textrm{s}i} \mid\mu_i, \sigma_i^2)$.
		The normal distribution and its mean $\mu_i$ are illustrated in blue.
		In the case of local optimization, only two discrete probability values that are closest to the current estimate of the contour distance $d_{\textrm{s}i}(\pmb{\theta})$ are considered.
		The two discrete probability values $p(d_{\textrm{s}i}^-\mid\omega_{\textrm{s}i},\pmb{l}_{\textrm{s}i})$ and $p(d_{\textrm{s}i}^+\mid\omega_{\textrm{s}i},\pmb{l}_{\textrm{s}i})$, which are used to approximate the first-order derivative, are colored in red.
	}\label{fig:t40}
\end{figure}

In the case of global optimization, the posterior probability distribution of individual correspondence lines is approximated by a normal distribution $\mathcal{N}(d_{\textrm{s}i}\mid\mu_i, \sigma_i^2)$.
The required mean and standard deviation $\mu_i$ and $\sigma_i$ are thereby estimated from a set of discretized contour distances $d_{\textrm{s}i}$ and their corresponding probability values.
An example of the approximation of a discrete posterior probability distribution is shown in Fig.~\ref{fig:t40}.
Based on the normal distribution, the first- and second-order derivatives are calculated as
\begin{gather}\label{eq:t44}
\frac{\partial\ln\big(p(d_{\textrm{s}i}\mid\omega_{\textrm{s}i},\pmb{l}_{\textrm{s}i})\big)}{\partial {d_{\textrm{s}i}}} \approx -\frac{1}{\sigma_i^2}(d_{\textrm{s}i} - \mu_i),\\ \label{eq:t44_0}
\frac{\partial^2\ln\big(p(d_{\textrm{s}i}\mid\omega_{\textrm{s}i},\pmb{l}_{\textrm{s}i})\big)}{\partial {d_{\textrm{s}i}}^2} \approx -\frac{1}{\sigma_i^2}.
\end{gather}
The approximated derivatives direct the optimization towards the mean $\mu_i$, using the variance $\sigma_i^2$ to consider uncertainty.
Note that while in the real world, the mean does not exactly coincide with the maximum, it is typically quite close.
At the same time, using the approximation has the advantage of fast convergence and that the optimization avoids local minima resulting from invalid pixel-wise posteriors and image noise.

Once the optimization is closer to the maximum, the global mean is not a good enough estimate, and more detailed refinement is required.
In such cases, the algorithm switches to local optimization.
We thereby use the probability values of the two discrete contour distances $d_{\textrm{s}i}^-$ and $d_{\textrm{s}i}^+$ that are closest to the current estimate $d_{\textrm{s}i}(\pmb{\theta})$ and approximate the first-order derivatives using a weighting term $\frac{\alpha_\textrm{s}}{\sigma_i^2}$ and finite differences 
\begin{equation}\label{eq:t45}
	\frac{\partial\ln\big(p(d_{\textrm{s}i}\mid\omega_{\textrm{s}i},\pmb{l}_{\textrm{s}i})\big)}{\partial {d_{\textrm{s}i}}} \approx \frac{\alpha_\textrm{s}}{\sigma_i^2}\ln\bigg(\frac{p(d_{\textrm{s}i}^+\mid\omega_{\textrm{s}i},\pmb{l}_{\textrm{s}i})}{p(d_{\textrm{s}i}^-\mid\omega_{\textrm{s}i},\pmb{l}_{\textrm{s}i})}\bigg).
\end{equation}
For second-order derivatives, the global approximation from Eq.~(\ref{eq:t44_0}) is used.
Note that weighting the first-order derivative with the variance $\sigma_i^2$ improves robustness because correspondence lines with high uncertainty are considered less important.
Simultaneously, the step size $\alpha_\textrm{s}$ helps to balance the weight and specifies how far the optimization proceeds, directly scaling the variation vector $\pmb{\hat{\theta}}$.
The same first- and second-order derivatives can also be derived using inverse-variance weighting and a constant curvature of $\frac{1}{\alpha_\textrm{s}}$ for the second-order derivative.
A detailed derivation of this interpretation is given in Appendix~\ref{sec:ad}.

Finally, apart from the choice of derivatives, the parameterization of smoothed step functions and the corresponding shape of posterior probability distributions significantly influences the optimization.
To study this effect, we consider the first-order derivatives of the log-posteriors that are shown in Fig.~\ref{fig:p60}.
While for Gaussian distributions, linear first-order derivatives lead to the estimation of the weighted mean over all correspondence lines, for Laplace distributions, binary derivatives guide the optimization towards the weighted median.
Note that this again corresponds well to the interpretation of local and global uncertainty modeled by the slope parameter $s_\textrm{h}$ and the amplitude parameter $\alpha_\textrm{h}$.
If only local uncertainty exists, it is advantageous to consider the magnitude of errors in the contour distance and optimize for the mean.
At the same time, in the case of global noise, it is reasonable to only consider the direction of errors, and conduct the optimization with respect to the median.

\section{Implementation}\label{sec:i}
The following section provides implementation details for the developed algorithm.
We thereby start with the generation of the sparse viewpoint model and the calculation of color histograms.
This is followed by a description of the tracking process.
Finally, we explain how known occlusions can be considered.
All mentioned parameter values are carefully chosen to maximize tracking quality while not requiring unreasonable amounts of computation.
Note that the source code of \textit{SRT3D} is publicly available on \textit{GitHub}\footnote[1]{ https://github.com/DLR-RM/3DObjectTracking} to ensure reproducibility and to allow full reusability.

\subsection{Sparse Viewpoint Model}\label{ssec:i0}
For the sparse viewpoint model, $n_\textrm{v} = 2562$ different views are considered.
They are generated by subdividing the triangles of an icosahedron 4 times, resulting in an angle of approximately $4^\circ$ between neighboring views.
Virtual cameras that are used for the rendering are placed at a distance of $0.8\,\unit{m}$ to the object center.
For all views, the orientation vector ${}_\textrm{M}\pmb{v}$ and a constant number of $n_\textrm{c}=200$ model points ${}_\textrm{M}\pmb{X}_i$ and normal vectors ${}_\textrm{M}\pmb{N}_i$ are computed.
In addition, for each point and view, we also compute so-called continuous distances for the foreground and background.
Continuous distances thereby describe the distance from the 2D model point $\pmb{x}_i$ along the line defined by the normal vector $\pmb{n}_i$ for which the foreground and background are not interrupted by each other.
After their computation in the rendered image, they are converted and stored in meters.
The values are later used by the tracker to disable individual correspondence lines for which continuous distances are below a certain threshold, and the assumption that only a single transition between foreground and background is present in the correspondence line is not sufficiently fulfilled.

\subsection{Color Histograms}\label{ssec:i1}
For the estimation of the color probability distributions $p(\pmb{y}\mid m_\textrm{f})$ and $p(\pmb{y}\mid m_\textrm{b})$, color histograms are used.
Each dimension of the RGB color space is discretized by $32$ equidistant bins, leading to a total of $32768$ values.
The computation of the color histograms is started either from the current pose estimate or from an initial pose, provided, for example, by a 3D object detection pipeline.
Based on this pose, 3D model points and normal vectors are projected into the image using Eq.~(\ref{eq:t20}) and (\ref{eq:t20_0}).
After an offset of one pixel, the first $18$ pixels are considered in both the positive and negative direction of the normal vector.
Pixel colors along this line are assigned to either the foreground or background histogram, depending on which side of the projected model point they are.
Note that fewer than $18$ pixels are considered if a transition between foreground and background occurs within a shorter distance.
Also, in cases where the contour location is more uncertain, it is reasonable to use an offset larger than one pixel.

Due to motion or dynamic illumination, color statistics of both the foreground and background are continuously changing during tracking.
To take those changes into account while at the same time considering previous observations, we use online adaptation.
Based on \cite{Bibby2008}, we thereby update the histograms as follows
\begin{equation}\label{eq:i10}
	p_t(\pmb{y}\mid m_i) = \alpha_i p(\pmb{y}\mid m_i) + (1 - \alpha_i) p_{t - 1}(\pmb{y}\mid m_i),
\end{equation}
with $i \in \{\textrm{f}, \textrm{b}\}$ and $\alpha_\textrm{f} = 0.2$ and $\alpha_\textrm{b} = 0.2$ the learning rates for the foreground and background, respectively.
Note that $p(\pmb{y}\mid m_i)$ is the observed histogram, while $p_t(\pmb{y}\mid m_i)$ and $p_{t-1}(\pmb{y}\mid m_i)$ are the adapted histograms of the current and previous time step, respectively.
For initialization, we directly use the observed histograms instead of blending them with previous values.

\subsection{Tracking Process}\label{ssec:i2}
To start tracking, an initial pose is required, which is typically provided either from a 3D object detection pipeline or from dataset annotations.
Based on this pose and a corresponding camera image, the color histograms for the foreground and background are initialized.
After initialization, a tracking step is executed for each new image that is streamed from the camera.
An overview of all computation that is performed in a single tracking step is given in Algorithm~\ref{al:i20}.
\begin{algorithm}
	\caption{Tracking Step}\label{al:i20}
	\begin{algorithmic}[1]
		\State Update camera image
		\For {$i=1,2,\ldots,7$}
		\State \textbf{Optional:} Render occlusion mask
		\State Find closest view of the sparse viewpoint model
		\State Define correspondence lines in the image
		\State Compute discrete distributions $p(d_{\textrm{s}i}\mid\omega_{\textrm{s}i},\pmb{l}_{\textrm{s}i})$
		\For {$j=1,2$}
		\State Calculate gradient $\pmb{g}$ and Hessian $\pmb{H}$
		\State Estimate variation $\pmb{\hat{\theta}}$ and update pose ${}_\textrm{C}\pmb{T}_\textrm{M}$
		\EndFor
		\EndFor
		\State Update color histograms $p(\pmb{y}\mid m_\textrm{f})$ and $p(\pmb{y}\mid m_\textrm{b})$
	\end{algorithmic} 
\end{algorithm}

Starting from a new image and the previous pose estimate ${}_\textrm{C}\pmb{T}_\textrm{M}$, we first retrieve the closest view of the sparse viewpoint model.
Model points ${}_\textrm{M}\pmb{X}_i$ and normal vectors ${}_\textrm{M}\pmb{N}_i$ are then projected into the image plane to define correspondence lines.
After that, continuous distances from the sparse viewpoint model are used to reject correspondence lines with distances that are below $6$ segments.
For the remaining correspondence lines, the posterior probability distribution $p(d_{\textrm{s}i}\mid\omega_{\textrm{s}i},\pmb{l}_{\textrm{s}i})$ is evaluated at $12$ discrete values $d_{\textrm{s}i}\in\{-5.5,-4.5,\dots,5.5\}$.
In the calculation, we use $8$ precomputed values for the smoothed step functions $h_\textrm{f}$ and $h_\textrm{b}$, corresponding to $x\in\{-3.5, -2.5,\dots,3.5\}$.
Also, a minimal offset $\Delta r_i$ is chosen such that the line coordinates $r_i$ point to pixel centers while the scaled line coordinates $r_{\textrm{s}i}$ ensure matching values for $x = r_{\textrm{s}i} - d_{\textrm{s}i}$.
In our case, this means that $r_{\textrm{s}i} \in \mathbb{Z}$.
Having computed the distributions, two iterations of the regularized Newton optimization are executed.
For the first iteration, the global optimization is used to quickly converge towards a rough pose estimate.
In the second iteration, the local optimization is employed to refine this pose, using a step size of $\alpha_\textrm{s} = 1.3$.
As regularization parameters, we use $\lambda_\textrm{r}=5000$ and $\lambda_\textrm{t}=500000$.

To find the final pose, the process is repeated seven times.
We thereby choose larger scales of $s=5$ for the first iteration and $s=2$ for the second and third iterations.
In all other iterations, a scale of $s=1$ is adopted.
This choice has the effect that a large area with low resolution is considered in the beginning, while short lines with high resolution are used in later iterations.
An example of correspondence lines at different scales is shown in Fig.~\ref{fig:r20}.
Note that scale values typically depend on the area that needs to be covered by the tracker and the size of frame-to-frame pose differences.
Finally, having estimated the pose for the current image, the prediction is used to update the color histograms.
After that, the tracker waits for a new image to arrive.

\subsection{Occlusion Modeling}\label{ssec:i3}
While the algorithm is quite robust to unknown occlusions, tracking results can be further improved by explicitly considering known occlusions.
For this, an ID is assigned to each known object.
All objects are then rendered into a depth image and an image that contains object ID values.
Using a custom shader, we combine information from the two images and compute an occlusion mask that binary encodes in each pixel which objects are visible.
To consider uncertainty in the object pose, the shader evaluates a region with a radius of $4$ pixels and assigns the object ID with the smallest depth value to the center.
If only the background is present, all object IDs are considered visible.
In order to improve efficiency, a smaller image with a fourth of the camera resolution is used.
Finally, to reject occluded correspondence lines, the algorithm simply checks occlusion mask values at correspondence line centers.

\section{Evaluation}\label{sec:e}
In this section, we present an extensive evaluation of our approach, \textit{SRT3D}.
Both the \textit{\ac{RBOT}} dataset \citep{Tjaden2018} and the \textit{\ac{OPT}} dataset \citep{Wu2017} are used to compare our method to the current state of the art in region-based tracking.
We thereby evaluate the quality of the predicted pose as well as the speed of the algorithm.
Also, a detailed parameter analysis is conducted that assesses the importance of different settings.
Finally, we discuss essential design considerations and remaining limitations.
In addition to the content in this section, we provide real-world videos on our project site\footnote[1]{ https://rmc.dlr.de/rm/staff/manuel.stoiber/ijcv2021} that demonstrate the tracker's performance.

\subsection{\acs{RBOT} Dataset}\label{ssec:e0}
\begin{figure}[t]
	\centering
	\input{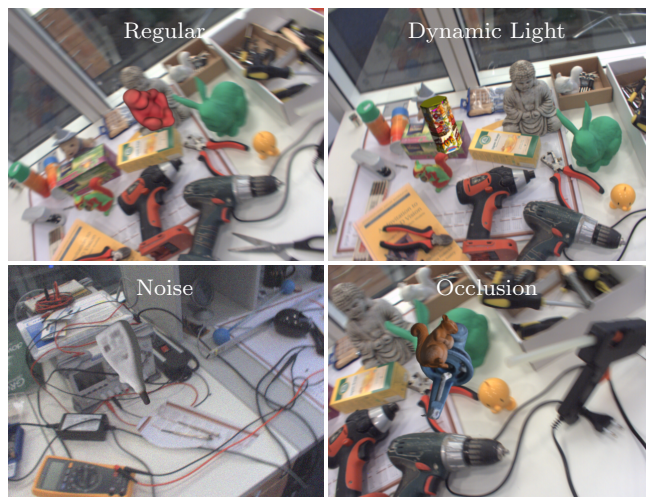}
	\caption{Images from the \textit{\ac{RBOT}} dataset \citep{Tjaden2018} with one example image for the \textit{regular}, \textit{dynamic light}, \textit{noise}, and \textit{occlusion} sequence.
	The sequences show the \textit{ape}, \textit{candy}, \textit{glue}, and \textit{vise} objects, respectively.
	In addition, the \textit{occlusion} sequence features a squirrel object that occludes the \textit{vise}.} \label{fig:e01}
\end{figure}

In the following, we first introduce the \textit{\ac{RBOT}} dataset, discuss the conducted experiments, and finally compare our results to the current state of the art.
The \textit{\ac{RBOT}} dataset consists of a collection of $18$ objects that are shown in Fig.~\ref{fig:e00}.
\begin{figure}[t]
	\centering
	\input{61_rbot_objects.tex}
	\caption{
		Overview of all objects in the \textit{\ac{RBOT}} dataset \citep{Tjaden2018}.
		Objects from the \textit{LINEMOD} dataset \citep{Hinterstoisser2013} and \textit{Rigid Pose} dataset \citep{Pauwels2013} are marked with $^\star$ and $^\diamond$, respectively.}\label{fig:e00}
\end{figure}
For each object, four sequences exist: a \textit{regular} version, one with \textit{dynamic light}, a sequence with both dynamic light and Gaussian \textit{noise}, and one with dynamic light and an additional squirrel object that leads to \textit{occlusion}.
An example image for each sequence is shown in Fig.~\ref{fig:e01}.\begin{table*}[h]
	\caption{Tracking success rates for state-of-the-art approaches on the \textit{\ac{RBOT}} dataset \citep{Tjaden2018}.
		Methods that are not purely region-based are indicated by a $^\star$.
		The best results are highlighted in bold.
		The second-best values are underlined.}
	\label{tab:e00}
	\begin{center}
\scriptsize
\begin{tabularx}{\textwidth}{@{\hspace{0.15cm}} l@{\hspace{-0.1cm}} *{17}{>{\centering\arraybackslash}X@{\hspace{-0.4cm}}} >{\centering\arraybackslash}X@{\hspace{-0.0cm}} c@{\hspace{0.15cm}}}
\hline
\noalign{\smallskip}
\textbf{Approach}&\rot{Ape} & \rot{Soda} & \rot{Vise} & \rot{Soup} & \rot{Camera} & \rot{Can} & \rot{Cat} & \rot{Clown} & \rot{Cube} & \rot{Driller} & \rot{Duck} & \rot{Egg Box} & \rot{Glue} & \rot{Iron} & \rot{Candy} & \rot{Lamp} & \rot{Phone} & \rot{Squirrel} &{\textbf{Avg.}}\\
\noalign{\smallskip}
\hline
\noalign{\medskip}

\noalign{Regular}
\noalign{\medskip}
\citet{Tjaden2018} & 85.0& 39.0& 98.9& 82.4& 79.7& 87.6& 95.9& 93.3& 78.1& 93.0& 86.8& 74.6& 38.9& 81.0& 46.8 & \underline{97.5}& 80.7& 99.4& 79.9\\
\citet{Zhong2020} & 88.8& 41.3& 94.0& 85.9& 86.9& 89.0& 98.5& 93.7& 83.1& 87.3& 86.2& 78.5& 58.6& 86.3& 57.9& 91.7& 85.0& 96.2& 82.7\\
\citet{Huang2020}$^\star$ & 91.9& 44.8 & \textbf{99.7}& 89.1& 89.3& 90.6& 97.4& 95.9& 83.9 & \underline{97.6}& 91.8& 84.4& 59.0 & \underline{92.5}& 74.3& 97.4& 86.4& 99.7& 86.9\\
\citet{Stoiber2020b}  & \underline{96.4}& 53.2& 98.8 & \underline{93.9}& 93.0& 92.7 & \underline{99.7} & \underline{97.1} & \underline{92.5}& 92.5 & \underline{93.7}& 88.5 & \underline{70.0}& 92.1 & \underline{78.8}& 95.5& 92.5& 99.6 & \underline{90.0}\\
\citet{Liu2021}$^\star$ & 93.7& 39.3& 98.4& 91.6& 84.6& 89.2& 97.9& 95.9& 86.3& 95.1& 93.4& 77.7& 61.5& 87.8& 65.0& 95.2& 85.7 & \textbf{99.8}& 85.5\\
\citet{Li2021}$^\star$ & 92.8& 42.6& 96.8& 87.5& 90.7& 86.2& 99.0& 96.9& 86.8& 94.6& 90.4& 87.0& 57.6& 88.7& 59.9& 96.5& 90.6& 99.5& 85.8\\
\citet{Sun2021}$^\star$ & 93.0 & \underline{55.2}& 99.3& 85.4 & \underline{96.1} & \underline{93.9}& 98.0& 95.6& 79.5 & \textbf{98.2}& 89.7 & \underline{89.1}& 66.5& 91.3& 60.6 & \textbf{98.6} & \underline{95.6}& 99.6& 88.1\\
SRT3D (Ours)  & \textbf{98.8} & \textbf{65.1} & \underline{99.6} & \textbf{96.0} & \textbf{98.0} & \textbf{96.5} & \textbf{100.0} & \textbf{98.4} & \textbf{94.1}& 96.9 & \textbf{98.0} & \textbf{95.3} & \textbf{79.3} & \textbf{96.0} & \textbf{90.3}& 97.4 & \textbf{96.2} & \textbf{99.8} & \textbf{94.2}\\
\hline
\noalign{\medskip}

\noalign{Dynamic Light}
\noalign{\medskip}
\citet{Tjaden2018} & 84.9& 42.0& 99.0& 81.3& 84.3& 88.9& 95.6& 92.5& 77.5& 94.6& 86.4& 77.3& 52.9& 77.9& 47.9& 96.9& 81.7& 99.3& 81.2\\
\citet{Zhong2020} & 89.7& 40.2& 92.7& 86.5& 86.6& 89.2& 98.3& 93.9& 81.8& 88.4& 83.9& 76.8& 55.3& 79.3& 54.7& 88.7& 81.0& 95.8& 81.3\\
\citet{Huang2020}$^\star$ & 91.8& 42.3& 98.9& 89.9& 91.3& 87.8& 97.6& 94.5& 84.5 & \underline{98.1}& 91.9& 86.7& 66.2& 90.9& 73.2 & \underline{97.1}& 89.2& 99.6& 87.3\\
\citet{Stoiber2020b}  & \underline{96.5}& 54.6& 99.1 & \underline{93.9}& 93.1 & \underline{94.7}& 99.5 & \underline{97.0} & \underline{93.0}& 93.4 & \underline{93.3} & \underline{92.6} & \underline{74.9} & \underline{91.0} & \underline{79.2}& 95.6& 89.8& 99.5 & \underline{90.6}\\
\citet{Liu2021}$^\star$ & 93.5& 38.2& 98.4& 88.8& 87.0& 88.5& 98.1& 94.4& 85.1& 95.1& 92.7& 76.1& 58.1& 79.6& 62.1& 93.2& 84.7& 99.6& 84.1\\
\citet{Li2021}$^\star$ & 93.5& 43.1& 96.6& 88.5& 92.8& 86.0 & \underline{99.6}& 95.5& 85.7& 96.8& 91.1& 90.2& 68.4& 86.8& 59.7& 96.1& 91.5& 99.2& 86.7\\
\citet{Sun2021}$^\star$ & 93.8 & \underline{55.9} & \textbf{99.6}& 85.6 & \textbf{97.7}& 93.7& 97.7& 96.5& 78.3 & \textbf{98.6}& 91.0& 91.6& 72.1& 90.7& 63.0 & \textbf{98.9} & \underline{94.4} & \textbf{100.0}& 88.8\\
SRT3D (Ours)  & \textbf{98.2} & \textbf{65.2} & \underline{99.2} & \textbf{95.6} & \underline{97.5} & \textbf{98.1} & \textbf{100.0} & \textbf{98.5} & \textbf{94.2}& 97.5 & \textbf{97.9} & \textbf{96.9} & \textbf{86.1} & \textbf{95.2} & \textbf{89.3}& 97.0 & \textbf{95.9} & \underline{99.9} & \textbf{94.6}\\
\hline
\noalign{\medskip}

\noalign{Noise}
\noalign{\medskip}
\citet{Tjaden2018} & 77.5& 44.5& 91.5& 82.9& 51.7& 38.4& 95.1& 69.2& 24.4& 64.3& 88.5& 11.2& 2.9& 46.7& 32.7& 57.3& 44.1& 96.6& 56.6\\
\citet{Zhong2020} & 79.3& 35.2& 82.6& 86.2& 65.1& 56.9& 96.9& 67.0& 37.5& 75.2& 85.4& 35.2& 18.9& 63.7& 35.4& 64.6& 66.3& 93.2& 63.6\\
\citet{Huang2020}$^\star$ & 89.0& 45.0& 89.5& 90.2& 68.9& 38.3& 95.9& 72.8& 20.1& 85.5& 92.2& 26.8& 15.8& 66.2& 52.2& 58.3& 65.1& 98.4& 65.0\\
\citet{Stoiber2020b} & 91.9& 53.3& 90.2 & \underline{92.6}& 67.9& 59.3& 98.4& 80.6 & \underline{43.5}& 78.1& 92.5& 44.0& 31.3& 72.3 & \underline{62.0}& 59.9& 71.7& 98.3& 71.5\\
\citet{Liu2021}$^\star$ & 84.7& 33.0& 88.8& 89.5& 56.4& 50.1& 94.1& 66.5& 32.3& 79.6 & \underline{94.2}& 29.6& 19.9& 63.4& 40.3& 61.6& 62.4& 96.9& 63.5\\
\citet{Li2021}$^\star$ & 89.1& 44.0& 91.6& 89.4& 75.2& 62.3 & \underline{98.6}& 77.3& 41.2& 81.5& 91.6& 54.5& 31.8& 65.0& 46.0 & \underline{78.5}& 69.6& 97.6& 71.4\\
\citet{Sun2021}$^\star$  & \underline{92.5} & \underline{56.2} & \textbf{98.0}& 85.1 & \textbf{91.7} & \textbf{79.0}& 97.7 & \underline{86.2}& 40.1 & \textbf{96.6}& 90.8 & \textbf{70.2} & \textbf{50.9} & \textbf{84.3}& 49.9 & \textbf{91.2} & \textbf{89.4} & \underline{99.4} & \underline{80.5}\\
SRT3D (Ours)  & \textbf{96.9} & \textbf{61.9} & \underline{95.4} & \textbf{95.7} & \underline{84.5} & \underline{73.9} & \textbf{99.9} & \textbf{90.3} & \textbf{62.2} & \underline{87.8} & \textbf{97.6} & \underline{62.2} & \underline{43.4} & \textbf{84.3} & \textbf{78.2}& 73.3 & \underline{83.1} & \textbf{99.7} & \textbf{81.7}\\
\hline
\noalign{\medskip}

\noalign{Unmodeled Occlusion}
\noalign{\medskip}
\citet{Tjaden2018} & 80.0& 42.7& 91.8& 73.5& 76.1& 81.7& 89.8& 82.6& 68.7& 86.7& 80.5& 67.0& 46.6& 64.0& 43.6& 88.8& 68.6& 86.2& 73.3\\
\citet{Zhong2020} & 83.9& 38.1& 92.4& 81.5& 81.3& 85.5 & \underline{97.5}& 88.9& 76.1& 87.5& 81.7& 72.7& 52.5& 77.2& 53.9& 88.5& 79.3& 92.5& 78.4\\
\citet{Huang2020}$^\star$ & 86.2& 46.3 & \underline{97.8}& 87.5& 86.5& 86.3& 95.7& 90.7& 78.8& 96.5& 86.0& 80.6& 59.9& 86.8& 69.6& 93.3& 81.8& 95.8& 83.6\\
\citet{Stoiber2020b} & 90.8& 51.7& 95.9 & \underline{88.5}& 88.0 & \underline{90.5}& 96.9& 91.6 & \underline{87.1}& 90.3& 86.4 & \underline{85.6}& 65.8& 87.0 & \underline{72.7}& 91.2& 84.0& 97.0 & \underline{85.6}\\
\citet{Liu2021}$^\star$ & 87.1& 36.7& 91.7& 78.8& 79.2& 82.5& 92.8& 86.1& 78.0& 90.2& 83.4& 72.0& 52.3& 72.8& 55.9& 86.9& 77.8& 93.0& 77.6\\
\citet{Li2021}$^\star$ & 89.3& 43.3& 92.2& 83.1& 84.1& 79.0& 94.5& 88.6& 76.2& 90.4 & \underline{87.0}& 80.7& 61.6& 75.3& 53.1& 91.1& 81.9& 93.4& 80.3\\
\citet{Sun2021}$^\star$  & \underline{91.3} & \underline{56.7} & \underline{97.8}& 82.0 & \underline{92.8}& 89.9& 96.6 & \underline{92.2}& 71.8 & \textbf{97.0}& 85.0& 84.6 & \underline{66.9} & \underline{87.7}& 56.1 & \underline{95.1} & \underline{89.8} & \underline{98.2}& 85.1\\
SRT3D (Ours)  & \textbf{96.5} & \textbf{66.8} & \textbf{99.0} & \textbf{95.8} & \textbf{95.0} & \textbf{95.9} & \textbf{100.0} & \textbf{97.6} & \textbf{92.2} & \underline{96.6} & \textbf{95.0} & \textbf{94.4} & \textbf{79.0} & \textbf{94.7} & \textbf{89.8} & \textbf{95.7} & \textbf{93.6} & \textbf{99.6} & \textbf{93.2}\\
\hline
\noalign{\medskip}

\noalign{Modeled Occlusion}
\noalign{\medskip}
\citet{Tjaden2018} & 82.0& 42.0& 95.7& 81.1& 78.7& 83.4& 92.8& 87.9& 74.3& 91.7& 84.8& 71.0& 49.1& 73.0& 46.3& 90.9& 76.2& 96.9& 77.7\\
\citet{Huang2020}$^\star$ & 87.8& 45.5 & \underline{98.1}& 87.2& 89.0& 89.8& 95.1& 91.4& 77.4 & \textbf{97.1}& 87.7& 83.0& 62.5& 88.6& 69.7 & \underline{94.1}& 86.0& 98.9& 84.9\\
\citet{Stoiber2020b}  & \underline{95.0} & \underline{53.8}& 97.8 & \underline{92.4} & \underline{90.6} & \underline{93.5} & \underline{99.1} & \underline{96.3} & \underline{91.5}& 92.6 & \underline{90.9} & \underline{91.3} & \underline{70.5} & \underline{91.8} & \underline{77.2}& 93.7 & \underline{87.0} & \underline{99.0} & \underline{89.1}\\
SRT3D (Ours)  & \textbf{97.9} & \textbf{68.3} & \textbf{99.2} & \textbf{95.4} & \textbf{96.8} & \textbf{96.4} & \textbf{99.6} & \textbf{98.6} & \textbf{93.0} & \underline{96.4} & \textbf{96.6} & \textbf{96.2} & \textbf{82.9} & \textbf{95.1} & \textbf{91.0} & \textbf{96.0} & \textbf{94.5} & \textbf{99.6} & \textbf{94.1}\\
\hline
\end{tabularx}
\end{center}

\end{table*}
Each sequence consists of 1001 semi-synthetic monocular images, where objects were rendered into real-world images, recorded from a hand-held camera that moves around a cluttered desk.

In the evaluation, experiments are performed as defined by \citet{Tjaden2018}.
The required translational and rotational errors are calculated as
\begin{equation}\label{eq:e00}
	e_{\textrm{t}}(t_k) = \big\lVert{}_\textrm{C}\pmb{t}_\textrm{M}(t_k) - {}_\textrm{C}\pmb{t}_{\textrm{M}_\textrm{gt}}(t_k) \big\rVert_2,
\end{equation}
\begin{equation}\label{eq:e01}
	e_{\textrm{r}}(t_k) = \cos^{-1}\hspace{-2pt}\bigg(\frac{\trace({}_\textrm{C}\pmb{R}_\textrm{M}(t_k)^\top\hspace{-1pt} {}_\textrm{C}\pmb{R}_{\textrm{M}_\textrm{gt}}(t_k)) - 1}{2}\bigg),\hspace{-1pt}
\end{equation}
where ${}_\textrm{C}\pmb{R}_{\textrm{M}_\textrm{gt}}(t_k)$ and ${}_\textrm{C}\pmb{t}_{\textrm{M}_\textrm{gt}}(t_k)$ are the ground-truth rotation matrix and translation vector for the frame $k \in \{0, \dots, 1000\}$.
A pose is considered successful if both $e_{\textrm{t}}(t_k) < 5\,\unit{cm}$ and  $e_{\textrm{r}}(t_k) < 5^\circ$.
After the initialization of the tracker with the ground-truth pose at $t_0$, the tracker runs until either the recorded sequence ends or tracking was unsuccessful.
In the case of unsuccessful tracking, the algorithm is re-initialized with the ground-truth pose at $t_k$.
For the \textit{occlusion} sequence, the method is evaluated with and without occlusion modeling.
In the case of occlusion modeling, both objects are tracked simultaneously.
Unsuccessful tracking of the occluding squirrel object is not considered in the reported tracking success.
Finally, for the remaining tracker settings, we use $\alpha_\textrm{h}=0.36$ and $s_\textrm{h} \to 0 $.
A detailed analysis of this choice is given in Sect.~\ref{ssec:e2}.

Results of the evaluation are shown in Table~\ref{tab:e00}.
Our approach is compared to the current state of the art in region-based tracking, as well as the edge-based methods of \citet{Huang2020}, algorithms of \citet{Li2021} and \citet{Sun2021} that combine edge and region information, and the method of \citet{Liu2021} that uses descriptor fields in addition to region-based techniques.
The comparison shows that \textit{SRT3D} performs significantly better than previous methods, achieving superior results for most objects and performing best on average.
This difference becomes even larger for purely region-based methods, with our algorithm performing best for almost all objects and sequences.
Considering the average success rate, our approach performs about five percentage points better than the combined method of \citet{Sun2021}, six percentage points better than \citet{Stoiber2020b}, nine percentage points better than \citet{Li2021}, and more than 14 percentage points better than the next best, dense, region-based approach by \citet{Zhong2019}.
The superior tracking success compared to our previous approach is especially interesting since the main differences are only an extended smoothed step function and some changes with respect to optimization.
Also, in comparison to all other, dense approaches, no advanced segmentation model is used, which, in theory, is a significant disadvantage.

In addition to tracking success, we also compare average runtimes.
A summary for the different algorithms is given in Table~\ref{tab:e01}.
\begin{table}
	\caption{Average runtimes per frame and usage of a GPU for state-of-the-art approaches.
	Methods that are not purely region-based are indicated by a $^\star$.
	For the occlusion modeling scenario, which considers the tracking of two objects, values are shown in parenthesis.}
	\label{tab:e01}
	\begin{center}
	\begin{tabularx}{8.4cm}{l >{\centering\arraybackslash}X c}
		\hline
		\noalign{\smallskip}
		\noalign{\smallskip}
		\textbf{Approach} & No GPU & Runtime\\
		\noalign{\smallskip}
		\hline
		\noalign{\smallskip}
		\citet{Tjaden2018} & \xmark & $15.5\sim21.8\,\unit{ms}$\\
		\citet{Zhong2020} & \xmark & $41.2\,\unit{ms}$\\
		\citet{Huang2020}$^\star$ & \xmark & $33.1\,\unit{ms}$\\
		\citet{Stoiber2020b} & \cmark (\xmark) & $1.0\,\unit{ms}$ ($7.4\,\unit{ms}$)\\
		\citet{Liu2021}$^\star$ & \xmark & $6.9\,\unit{ms}$\\
		\citet{Li2021}$^\star$ & \xmark & $32.1\,\unit{ms}$\\
		\citet{Sun2021}$^\star$ & \xmark & $40.0\sim50.0\,\unit{ms}$\\
		SRT3D (Ours) & \cmark (\xmark) & $1.1\,\unit{ms}$ ($5.1\,\unit{ms}$)\\
		\noalign{\smallskip}
		\hline
	\end{tabularx}
\end{center}

\end{table}
The evaluation of \textit{SRT3D} and our previous method was conducted on the same computer with an \textit{Intel Xeon E5-1630 v4} CPU and a \textit{Nvidia Quadro P600} GPU.
Because of the similarities of the two approaches, we obtain a comparable average runtime of $1.1\,\unit{ms}$ for the case without occlusion modeling and an improved average execution time of $5.1\,\unit{ms}$ for the \textit{modeled occlusion} scenario.
Note that in the case of occlusion modeling, occlusion masks have to be rendered, and the reported time is for the simultaneous tracking of two objects.
In comparison, except for the algorithm of \citet{Liu2021}, for which the execution time is six times higher, all other methods report average runtimes that are more than one order of magnitude larger.
The difference is even more impressive since \textit{SRT3D} and our previous approach only utilize a single CPU core and do not require a GPU.
In contrast, most competing methods typically use multithreading and heavily depend on a GPU.
In conclusion, while different resources and computers were used, the obtained results highlight the superior efficiency of our sparse region-based method.

\subsection{\acs{OPT} Dataset}\label{ssec:e1}
\begin{figure}[t]
	\centering
	\input{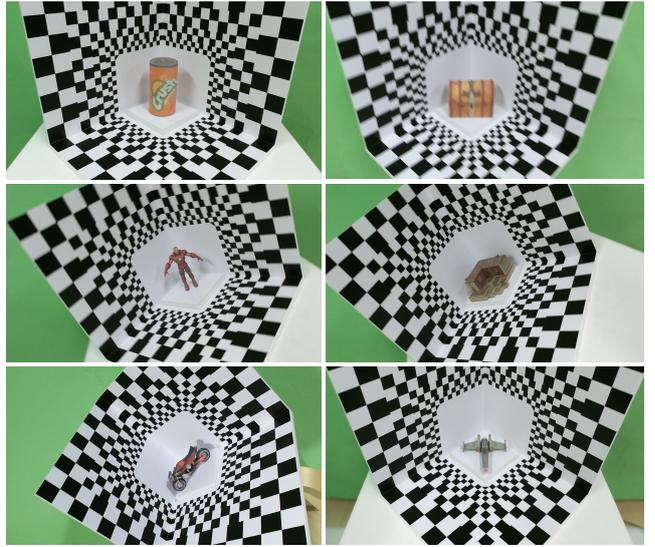}
	\caption{
		Images from the \textit{\ac{OPT}} dataset \citep{Wu2017}, featuring the \textit{soda}, \textit{chest}, \textit{ironman}, \textit{house}, \textit{bike}, and \textit{jet} object.
	} \label{fig:e10}
\end{figure}

While the semi-synthetic \textit{\ac{RBOT}} dataset features a large number of objects, a difficult, highly cluttered background, and perfect ground-truth, objects are simulated with limited realism, and only very little motion blur is applied.
Those shortcomings are complemented by the \textit{\ac{OPT}} dataset \citep{Wu2017}, which contains real-world recordings of 3D printed objects on a white background with different speeds and levels of motion blur.
In total, the dataset includes six objects and consists of 552 real-world sequences with various lighting conditions and defined trajectories recorded by a robot arm.
An example image for each object is shown in Fig.~\ref{fig:e10}.
The sequences are classified into the following categories: translation, forward and backward, in-plane rotation, out-of-plane rotation, flashing light, moving light, and free motion.

In the experiments, the metric of \citet{Wu2017} is used.
For this, we compute the average vertex error
\begin{equation}
	e_\textrm{v}(t_k) = \frac{1}{n}\sum_{i=1}^n\big\lVert \big({}_\textrm{M}\pmb{\widetilde{X}}_i -  {}_\textrm{M}\pmb{T}_{\textrm{M}_\textrm{gt}}(t_k){}_\textrm{M}\pmb{\widetilde{X}}_i\big)_{3\times 1}\big\rVert_2,
\end{equation}
with $\pmb{\widetilde{X}}_i$ a vertex in the 3D mesh geometry of the object and $n$ the number of vertices.
Tracking is considered successful if $e_\textrm{v}(t_k) < k_\textrm{e}d$, where $d$ is the object diameter computed from the maximum vertex distance and $k_\textrm{e}$ is an error threshold.
The tracking quality for all frames is then measured using an \ac{AUC} score that integrates the percentage value of successfully tracked poses over the interval $k_\textrm{e} \in [0,0.2]$.
This results in \ac{AUC} scores between zero and twenty.
For the tracker, the amplitude parameter $\alpha_\textrm{h}=0.42$ and the slope parameter $s_\textrm{h} = 0.5$ are used.
Also, for the rotationally symmetric \textit{soda} object, a larger rotational regularization parameter of $\lambda_\textrm{r}=500000$ is adopted.
The main reason is that the object geometry of the \textit{soda} object does not constrain the rotation around the vertical axis.
In such cases, fluctuations in the gradient and Hessian can lead to drift in the object's orientation.
Using more regularization allows us to mitigate this problem.

Results for the experiments on the \textit{\ac{OPT}} dataset are shown in Table.~\ref{tab:e10}.
\begin{table}
	\caption{\ac{AUC} scores between zero and twenty for the evaluation on the \textit{\ac{OPT}} dataset  \citep{Wu2017}, comparing our approach to multiple other algorithms.
	The best results are highlighted in bold.
	The second-best values are underlined.}
	\label{tab:e10}
	\begin{center}
\scriptsize
\begin{tabularx}{8.4cm}{@{\hspace{0.15cm}} l@{\hspace{-0.15cm}} *{5}{>{\centering\arraybackslash}X@{\hspace{-0.4cm}}} >{\centering\arraybackslash}X@{\hspace{-0.1cm}} c@{\hspace{0.15cm}}}
\hline
\noalign{\smallskip}
\textbf{Approach}&\rot{Soda} & \rot{Chest} & \rot{Ironman} & \rot{House} & \rot{Bike} & \rot{Jet} &\textbf{Avg.}\\
\noalign{\smallskip}
\hline
\noalign{\smallskip}
PWP3D &5.87&5.55&3.92&3.58&5.36&5.81&5.01\\
ElasticFusion&1.90&1.53&1.69&2.70&1.57&1.86&1.87\\
UDP&8.49&6.79&5.25&5.97&6.10&2.34&5.82\\
ORB-SLAM2&13.44&\underline{15.53}&11.20&\textbf{17.28}&10.41&9.93&12.97\\
\cite{Bugaev2018}&\underline{14.85}&14.97&\underline{14.71}&14.48&12.55&\textbf{17.17}&\underline{14.79}\\
\cite{Tjaden2018}&8.86&11.76&11.99&10.15&11.90&13.22&11.31\\
\cite{Zhong2020}&9.01&12.24&11.21&13.61&12.83&15.44&12.39\\
\cite{Li2021}&9.00&14.92&13.44&13.60&\underline{12.85}&10.64&12.41\\
SRT3D (Ours)&\textbf{15.64}&\textbf{16.30}&\textbf{17.41}&\underline{16.36}&\textbf{13.02}&\underline{15.64}&\textbf{15.73}\\
\noalign{\smallskip}
\hline
\end{tabularx}
\end{center}

\end{table}
We thereby compare \textit{SRT3D} to state-of-the-art region-based tracking approaches, as well as an approach from \citet{Bugaev2018} and prominent methods such as \textit{PWP3D} \citep{Prisacariu2012}, \textit{ElasticFusion} \citep{Whelan2015}, \textit{UDP} \citep{Brachmann2016}, and \textit{ORB-SLAM2} \citep{MurArtal2017}.
Note that not all algorithms are dedicated 3D tracking solutions.
\textit{UDP} is a monocular pose detection method, while \textit{ElasticFusion} and \textit{ORB-SLAM2} are visual SLAM approaches for camera pose localization that are applied to the silhouette of the object.
For more information on the evaluation of those algorithms, please refer to \citet{Wu2017}.

The comparison shows that our approach performs significantly better than the current state of the art in region-based tracking developed by \citet{Li2021}, \citet{Zhong2020} and \citet{Tjaden2018}, achieving higher \ac{AUC} scores for each of the six objects.
Also, compared to none region-based approaches, we are able to report the highest score for four out of six objects and perform best on average.
This is even more remarkable since \textit{ORB-SLAM2}, which reports better results for the \textit{house} object, uses gradient-based corner features.
In contrast to \textit{SRT3D}, the algorithm is thus not constrained to the contour but considers information over the entire silhouette.
Also, the edge-based algorithm of \citet{Bugaev2018}, which performs best for the \textit{jet} object, uses basin-hopping for global optimization, and, with an average reported runtime of $683\,\unit{ms}$, is not real-time capable.
In conclusion, the obtained results demonstrate that the excellent performance of \textit{SRT3D} on simulated data translates well to applications in the real world.

\subsection{Parameter Analysis}\label{ssec:e2}
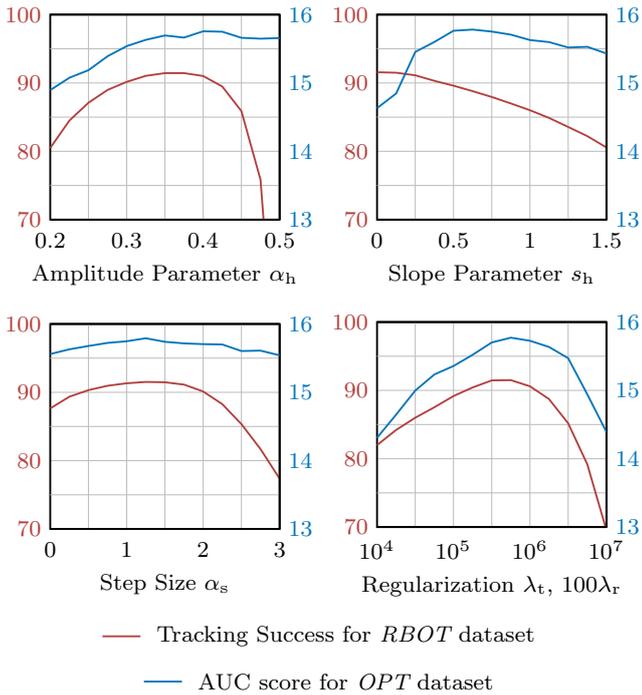
\begin{figure}[t]
	\centering
	\begin{tikzpicture}
	\clip (-4.15,-5.2) rectangle + (8.4, 9.2);
	\small
	\newcommand\plotwidth{4.6cm};
	\newcommand\plothight{4.3cm};
	\newcommand\xdistance{2.15cm}
	\newcommand\ydistance{2.05cm}
	
	\node at (-\xdistance, \ydistance) {
	\begin{tikzpicture}
	\begin{axis}[
		line width=0.7pt,
		width=\plotwidth,
		height=\plothight,
		xlabel near ticks,
		xlabel={Amplitude Parameter $\alpha_\textrm{h}$},
		xmin=0.2, xmax=0.5,
		ymin=70, ymax=100,
		xtick={0.2,0.3,0.4,0.5},
		xticklabel style={yshift=-.2em},
		extra x ticks={0.25,0.35,0.45},
		extra x tick style={xticklabels={}},
		ytick={70,80,90,100},
		yticklabel style=dlrred,
		extra y ticks={75,85,95},
		extra y tick style={yticklabels={}},
		tick style={draw=none},
		grid=both]
		\addplot[color=dlrred] coordinates {
			(0.2,80.48)(0.225,84.49)(0.25,87.11)(0.275,88.98)(0.3,90.16)(0.325,91.04)(0.35,91.44)(0.375,91.44)(0.4,91.02)(0.425,89.48)(0.45,85.88)(0.475,75.84)(0.49,53.8)
		};
	\end{axis}
	\begin{axis}[
		line width=0.7pt,
		width=\plotwidth,
		height=\plothight,
		xmin=0.2, xmax=0.5,
		ymin=13, ymax=16,
		xmajorticks=false,
		ytick={13,14,15,16},
		yticklabel style=dlrblue,
		tick style={draw=none},
		axis y line*=right,]
		\addplot[color=dlrblue]
		coordinates {
			(0.2,14.895)(0.225,15.076)(0.25,15.187)(0.275,15.391)(0.3,15.539)(0.325,15.631)(0.35,15.695)(0.375,15.664)(0.4,15.756)(0.425,15.75)(0.45,15.661)(0.475,15.648)(0.5,15.657)
		};
	\end{axis}
	\end{tikzpicture}};

	\node at (\xdistance, \ydistance) {
	\begin{tikzpicture}
	\begin{axis}[
		line width=0.7pt,
		width=\plotwidth,
		height=\plothight,
		xlabel near ticks,
		xlabel={Slope Parameter $s_\textrm{h}$},
		xmin=0, xmax=1.5,
		ymin=70, ymax=100,
		xtick={0,0.5,1.0,1.5},
		xticklabel style={yshift=-.2em},
		extra x ticks={0.25,0.75,1.25},
		extra x tick style={xticklabels={}},
		ytick={70,80,90,100},
		yticklabel style=dlrred,
		extra y ticks={75,85,95},
		extra y tick style={yticklabels={}},
		tick style={draw=none},
		grid=both]
		\addplot[color=dlrred] coordinates {
			(0.0,91.57)(0.125,91.50)(0.25,91.12)(0.375,90.31)(0.5,89.60)(0.625,88.80)(0.75,87.95)(0.875,87.00)(1.0,86.01)(1.125,84.88)(1.25,83.56)(1.375,82.22)(1.5,80.56)
			};
	\end{axis}
	\begin{axis}[
		line width=0.7pt,
		width=\plotwidth,
		height=\plothight,
		xmin=0, xmax=1.5,
		ymin=13, ymax=16,
		xmajorticks=false,
		ytick={13,14,15,16},
		yticklabel style=dlrblue,
		tick style={draw=none},
		axis y line*=right,]
		\addplot[color=dlrblue]
		coordinates {
			(0.0,14.631)(0.125,14.846)(0.25,15.454)(0.375,15.600)(0.5,15.764)(0.625,15.781)(0.75,15.752)(0.875,15.707)(1.0,15.628)(1.125,15.598)(1.25,15.519)(1.375,15.529)(1.5,15.431)
		};
	\end{axis}
	\end{tikzpicture}};

	\node at (-\xdistance, -\ydistance) {
	\begin{tikzpicture}
	\begin{axis}[
		line width=0.7pt,
		width=\plotwidth,
		height=\plothight,
		xlabel near ticks,
		xlabel={Step Size $\alpha_\textrm{s}$},
		xmin=0, xmax=3.0,
		ymin=70, ymax=100,
		xtick={0,1.0,2,3},
		xticklabel style={yshift=-.2em},
		extra x ticks={0.5,1.5,2.5},
		extra x tick style={xticklabels={}},
		ytick={70,80,90,100},
		yticklabel style=dlrred,
		extra y ticks={75,85,95},
		extra y tick style={yticklabels={}},
		tick style={draw=none},
		grid=both]
		\addplot[color=dlrred] coordinates {
			(0,87.64)(0.25,89.36)(0.5,90.32)(0.75,90.96)(1.0,91.32)(1.25,91.52)(1.5,91.47)(1.75,91.13)(2.0,90.12)(2.25,88.27)(2.5,85.35)(2.75,81.69)(3.0,77.36)
		};
	\end{axis}
	\begin{axis}[
		line width=0.7pt,
		width=\plotwidth,
		height=\plothight,
		xmin=0, xmax=3.0,
		ymin=13, ymax=16,
		xmajorticks=false,
		ytick={13,14,15,16},
		yticklabel style=dlrblue,
		tick style={draw=none},
		axis y line*=right,]
		\addplot[color=dlrblue]
		coordinates {
			(0,15.560)(0.25,15.629)(0.5,15.678)(0.75,15.723)(1.0,15.747)(1.25,15.791)(1.5,15.739)(1.75,15.716)(2.0,15.705)(2.25,15.700)(2.5,15.604)(2.75,15.610)(3.0,15.541)
		};
	\end{axis}
	\end{tikzpicture}};

	\node at (\xdistance, -\ydistance) {
	\begin{tikzpicture}
	\begin{axis}[
		xmode=log,
		line width=0.7pt,
		width=\plotwidth,
		height=\plothight,
		xlabel near ticks,
		xlabel={Regularization  $ \lambda_\textrm{t}$, $100\lambda_\textrm{r}$},
		xmin=1e4, xmax=1e7,
		ymin=70, ymax=100,
		xtick={1e4,1e5,1e6,1e7, 2e8},
		xticklabel style={yshift=-.2em},
		extra x ticks={31600,316000,3160000},
		extra x tick style={xticklabels={}},
		ytick={70,80,90,100},
		yticklabel style=dlrred,
		extra y ticks={75,85,95},
		extra y tick style={yticklabels={}},
		tick style={draw=none},
		grid=both]
		\addplot[color=dlrred] coordinates {
			(10000,81.99)(17700,84.20)(31600,86.00)(56200,87.53)(100000,89.17)(177000,90.42)(316000,91.46)(562000,91.50)(1000000,90.60)(1770000,88.74)(3160000,85.18)(5620000,79.18)(10000000,69.54)
		};
	\end{axis}
	\begin{axis}[
		xmode=log,
		line width=0.7pt,
		width=\plotwidth,
		height=\plothight,
		xmin=1e4, xmax=1e7,
		ymin=13, ymax=16,
		xmajorticks=false,
		ytick={13,14,15,16},
		yticklabel style=dlrblue,
		tick style={draw=none},
		axis y line*=right,]
		\addplot[color=dlrblue]
		coordinates {
			(10000,14.307)(17700,14.643)(31600,14.997)(56200,15.233)(100000,15.357)(177000,15.519)(316000,15.701)(562000,15.771)(1000000,15.726)(1770000,15.636)(3160000,15.471)(5620000,14.942)(10000000,14.391)
		};
	\end{axis}
	\end{tikzpicture}};

	\node at (0, -4.45) {
	\begin{tikzpicture}
		\draw [dlrred, line width=0.7pt] (-0.5, 0) -- (0, 0);
		\node at (0.1, 0)[anchor = west] {Tracking Success for \textit{\ac{RBOT}} dataset};
	\end{tikzpicture}};
	\node at (0, -5.05) {
	\begin{tikzpicture}
		\draw [dlrblue, line width=0.7pt] (-0.5, 0) -- (0, 0);
		\node at (0.1, 0)[anchor = west] {\ac{AUC} score for \textit{\ac{OPT}} dataset};
	\end{tikzpicture}};
\end{tikzpicture}
	\caption{
		Average tracking success for the \textit{\ac{RBOT}} dataset and average \ac{AUC} score for the \textit{\ac{OPT}} dataset over different values of the amplitude parameter $\alpha_\textrm{h}$, slope parameter $s_\textrm{h}$, step size $\alpha_\textrm{s}$, and the rotational and translational regularization parameters $\lambda_\textrm{r}$ and $\lambda_\textrm{t}$.
		For the evaluation of the regularization parameters, we set $\lambda_\textrm{t} = 100 \lambda_\textrm{r}$.
	} \label{fig:e20}
\end{figure}

Having evaluated the performance of our approach, we want to foster our understanding of different parameter values.
For this, the average success rate for the \textit{\ac{RBOT}} dataset and the average \ac{AUC} score for the \textit{\ac{OPT}} dataset are plotted over different parameter values.
The plots are shown in Fig.~\ref{fig:e20}.
Note that the success rate and \ac{AUC} score are computed over all objects and sequences.
Except for the parameter that is analyzed, the same settings as in Sect.~\ref{ssec:e0} and Sect.~\ref{ssec:e1} are used.

The evaluation of the amplitude parameter $\alpha_\textrm{h}$ shows that while it significantly influences the tracking success, the effect on the \ac{AUC} score is much smaller.
Knowing that the amplitude parameter models a constant level of noise, this makes sense since the \textit{\ac{RBOT}} dataset features highly cluttered images while the \textit{\ac{OPT}} dataset only contains a constant white background.
For the slope parameter $s_\textrm{h}$, the highest tracking success is observed for $s_\textrm{h} \to 0$, and the best \ac{AUC} score is obtained at $s_\textrm{h} = 0.5$.
Again, this is well explained by the theoretical interpretation according to which the slope parameter models local uncertainty.
Given perfect information about the object geometry for the semi-synthetic \textit{\ac{RBOT}} dataset, we do not expect any local uncertainty.
At the same time, for the \textit{\ac{OPT}} dataset, with imperfectly 3D printed objects and recorded real-world images, it is important that a larger parameter is chosen that allows for a defined level of uncertainty.

Studying the plot of the step size $\alpha_\textrm{s}$, we observe a relatively large plateau around one, with maximum values at $\alpha_\textrm{s} = 1.3$ for both the tracking success and the \ac{AUC} score.
This suggests a low dependency between the parameter and different image data.
Particularly interesting are also the results for $\alpha_\textrm{s} = 0$.
For this setting, no local optimization is considered, showing the capability of the global optimization alone.
The good results highlight the excellent performance of the adopted global approximation.

Finally, for the evaluation of regularization, the rotational and translational parameters are modified simultaneously.
To consider the different units of radians and meters, we define $\lambda_\textrm{t} = 100 \lambda_\textrm{r}$.
Like in previous evaluations of the \textit{soda} object, we increase the rotational parameter and use $\lambda_\textrm{r} =  \lambda_\textrm{t}$.
The resulting plot of the tracking success and the \ac{AUC} score demonstrates the high importance of regularization.
If values are chosen too small, the optimization is unstable for directions in which no or very little information is available.
At the same time, if parameters are too large, the optimization is slowed down, and the final pose might not be reached.
It is thus important to find values that lie in between.
In our experience, a good approximation is to use regularization parameters that are in the same order of magnitude as the maximum rotational and translational diagonal elements of the Hessian matrix.

In conclusion, the parameter analysis demonstrates that theoretical interpretations from Sect.~\ref{sec:p} and Sect.~\ref{sec:t} correspond well to experimental results.
In addition to fostering our understanding, this explainability helps to guide the parameter search for new applications.
Moreover, the results in Fig.~\ref{fig:e20} demonstrate that all parameters are well-behaved, with large plateaus around the maximum and no sudden jumps.
This has the advantage that parameters are easy to tune, with a broad range of values achieving satisfying results.

\subsection{Discussion}\label{ssec:e3}
The conducted experiments demonstrate the excellent performance of \textit{SRT3D}.
In the following, we want to discuss design considerations that are essential in achieving those results and shed some light on the remaining limitations of the algorithm.
With respect to computational efficiency, the biggest performance gain is attributed to the correspondence line model and the sparse nature of the method.
In addition, the sparse viewpoint model provides a highly efficient representation, which requires only a simple search to obtain the object contour for the current pose. 
Also, in contrast to dense methods, it is not necessary to compute a 2D signed distance function, but one can simply use the contour distance.
Finally, the discrete scale-space formulation reduces the amount of computation further by combining multiple pixels into segments and supporting the use of precomputed smoothed step functions.

For the quality of the pose estimate, multiple aspects have to be considered.
The first important factor is the use of smoothed step functions that provide a realistic modeling of local and global uncertainty.
Consequently, this leads to reliable posterior probability distributions.
Also, due to the one-dimensionality of correspondence lines and the discrete scale-space implementation, we are able to sample values over posterior probability distributions in reasonable time.
This allows us to calculate the mean and the variance.
Both estimates constitute the basis for fast-converging global optimization that is independent of local minima.
In addition, knowledge about the uncertainty of individual correspondence lines is also used for local optimization, where numerical first-order derivatives are weighted according to the inverse variance.
Finally, Tikhonov regularization is another important factor, which helps to constrain the estimate with respect to the previous pose, stabilizing the optimization for directions in which no or very little information is available.

While the described algorithm achieves remarkable results and works very well in a wide variety of applications, some challenges remain.
The main limitations are thereby very similar to other region-based methods.
The biggest constraint is that objects have to be rigid and that an accurate 3D model has to be known.
Also, the background has to be distinguishable from the object.
If large areas in the background contain colors that are also present in the object, the final result might be perturbed.
Another challenge comes from ambiguities where the object silhouette is very similar in the vicinity of a particular pose.
Naturally, in such cases, there is not enough information, and it is impossible for the algorithm to converge towards the correct pose.
Also, like most tracking approaches, the algorithm can only be used for local optimization with a limit to the maximum pose difference from one frame to the next.
Finally, if large parts of the object are occluded, the visible part of the contour might not fully constrain the pose of the object, leading to erroneous estimates.
To illustrate all the described failure cases, we provide a video on our project site\footnote[1]{ https://rmc.dlr.de/rm/staff/manuel.stoiber/ijcv2021}.

\section{Conclusion}
In this work, we proposed \textit{SRT3D}, a highly efficient, sparse approach to region-based 3D object tracking that uses correspondence lines to find the pose that best explains the segmentation of the image.
In addition to a thorough mathematical derivation of correspondence lines, a big contribution of this work is the development of smoothed step functions that allow the modeling of both local and global uncertainty.
The effects of this modeling were analyzed in detail with respect to both theoretical posterior probability distributions and the quality of the final tracking result.
For the maximization of the pose-dependent joint posterior probability, we proposed the use of an initial, global optimization towards the mean and a consecutive, local optimization that considers discrete distribution values.
We also developed a novel approximation for the local first-order derivative that weights the finite difference value with the inverse variance.
Finally, in multiple experiments on the \textit{\ac{RBOT}} and the \textit{\ac{OPT}} dataset, we demonstrated that our algorithm outperforms the current state of the art in region-based tracking by a considerable margin both in terms of quality and efficiency.

Thanks to this superior performance, we are confident that our approach is useful to a wide range of applications in robotics and augmented reality.
Because of its general formulation, it is easy to conceive ideas that extend the method.
One possible direction would be to include other developments in region-based tracking, such as advanced segmentation models or occlusion detection.
Also, it might be useful to consider additional information, like depth or texture.
Finally, we want to highlight that the developed correspondence line model is not limited to the context of 3D tracking but might also be useful to other applications.
One possible example is image segmentation.
Other methods might thereby show similar progress in terms of quality and efficiency, improving their applicability to the real world.

\appendixtitleon
\begin{appendices}

\section{Extended Probabilistic Model}\label{sec:aa}
In the following, we establish the relation between the smoothed step functions proposed in Sect.~\ref{ssec:p5} and an extended probabilistic model with $m \in \{m_\textrm{f}, m_\textrm{b}, m_\textrm{n}\}$.
For the derivation, we start from an extended definition of the pixel-wise posterior probability
\begin{equation} \label{eq:aa0}
	p(m_i \mid \pmb{y}) = \frac{p(\pmb{y}\mid m_i) p(m_i)}{\sum_{j\in\{\textrm{f},\textrm{b},\textrm{n}\}}p(\pmb{y}\mid m_j)p(m_j)} , \quad i\in\{\textrm{f}, \textrm{b}, \textrm{n}\},
\end{equation}
where, in contrast to Eq.~(\ref{eq:p33}), a noise model $m_\textrm{n}$ is considered in addition to the foreground and background model.
Using the parameter $\alpha_\textrm{h} \in [0, 0.5]$, the model probabilities are defined as
\begin{gather} \label{eq:aa1}
	p(m_\textrm{f}) = p(m_\textrm{b}) = \alpha_\textrm{h},\\
	p(m_\textrm{n}) = 1 - 2\alpha_\textrm{h}.
\end{gather}
For the conditional color probability given the noise model, the conditional probabilities with respect to the foreground and background are combined as follows
\begin{equation}\label{eq:aa2}
	p(\pmb{y}\mid m_\textrm{n}) = \frac{1}{2}\big(p(\pmb{y}\mid m_\textrm{f}) + p(\pmb{y}\mid m_\textrm{b})\big).
\end{equation}
Introducing the definitions from Eqs.~(\ref{eq:aa1}) to (\ref{eq:aa2}) into Eq.~(\ref{eq:aa0}) and performing some simplifications results in the following pixel-wise posterior probabilities for the foreground and background model
\begin{equation} \label{eq:aa3}
	p(m_i\mid \pmb{y}) = \frac{2\alpha_\textrm{h}p(\pmb{y}\mid m_i)}{p(\pmb{y}\mid m_\textrm{f}) + p(\pmb{y}\mid m_\textrm{b})} , \quad i\in\{\textrm{f}, \textrm{b}\}.
\end{equation}
Also, we obtain the following constant pixel-wise posterior probability for the noise model
\begin{equation} \label{eq:aa4}
	p(m_\textrm{n}\mid \pmb{y}) = p(m_\textrm{n}) = 1 - 2\alpha_\textrm{h}.
\end{equation}

Based on Eq.~(\ref{eq:p32}), the extended posterior probability can be calculated as follows
\begin{equation} \label{eq:aa6_0}
	p(d\mid r, \pmb{y}) \propto \sum_{i \in \{\textrm{f}, \textrm{b}, \textrm{n}\}}p(r\mid d, m_i) p(m_i\mid \pmb{y}).
\end{equation}
To abbreviate some of the terms in Eq.~(\ref{eq:aa3}), the following definition of pixel-wise posterior probabilities from Eq.~(\ref{eq:p34}) is introduced
\begin{equation}\label{eq:aa5}
		p_i(r) = \frac{p(\pmb{y}\mid m_i)}{p(\pmb{y}\mid m_\textrm{f}) + p(\pmb{y}\mid m_\textrm{b})} , \quad i\in\{\textrm{f}, \textrm{b}\}.
\end{equation}
We then use the derived pixel-wise posterior probabilities from Eq.~(\ref{eq:aa3}) and (\ref{eq:aa4}) together with the abbreviation from Eq.~(\ref{eq:aa5}) to write the posterior probability in Eq.~(\ref{eq:aa6_0}) as follows
\begin{equation} \label{eq:aa6}
	\begin{split}
	p(d\mid r, \pmb{y}) \propto\
	&2\alpha_\textrm{h} h_\textrm{f}(r-d) p_\textrm{f}(r)\, + \\
	&2\alpha_\textrm{h} h_\textrm{b}(r-d) p_\textrm{b}(r) + \frac{1}{2}(1 - 2\alpha_\textrm{h}),
	\end{split}
\end{equation}
where a constant probability $p(r\mid d,m_\textrm{n}) = \frac{1}{2}$ was used to model the indifference of the line coordinate $r$ given the noise model $m_\textrm{n}$, and where the smoothed step functions $h_\textrm{f}$ and $h_\textrm{b}$ model the line coordinate probabilities $p(r\mid d,m_\textrm{f})$ and $p(r\mid d,m_\textrm{b})$.
For the extension of Eq.~(\ref{eq:aa6}), we apply the following definitions
\begin{gather}\label{eq:aa7}
	h_\textrm{f}(x) = \frac{1}{2} - f(x),\\
	h_\textrm{b}(x) = \frac{1}{2} + f(x), 
\end{gather}
and the identity
\begin{equation}\label{eq:aa8}
	p_\textrm{f}(r) + p_\textrm{b}(r) = 1,
\end{equation}
to write
\begin{equation} \label{eq:aa9}
	\begin{split}
	p(d\mid r, \pmb{y}) \propto\
	&\alpha_\textrm{h}p_\textrm{f}(r) - 2 \alpha_\textrm{h}f(r-d)p_\textrm{f}(r)\,+\\ 
	&\alpha_\textrm{h}p_\textrm{b}(r) + 2 \alpha_\textrm{h}f(r-d)p_\textrm{b}(r)\,+\\
	&\frac{1}{2}\big(p_\textrm{f}(r) + p_\textrm{b}(r)\big)-\alpha_\textrm{h}\big(p_\textrm{f}(r) + p_\textrm{b}(r)\big).
	\end{split}\hspace{-1cm}
\end{equation}
This can then be simplified to
\begin{equation} \label{eq:aa10}
	\begin{split}
		p(d\mid r, \pmb{y}) \propto\
		&\bigg(\frac{1}{2} - 2 \alpha_\textrm{h}f(r-d)\bigg)p_\textrm{f}(r)\,+ \\ 
		&\bigg(\frac{1}{2} + 2 \alpha_\textrm{h}f(r-d)\bigg)p_\textrm{b}(r).
	\end{split}
\end{equation}
Finally, after introducing the slope function $f(x) = \frac{1}{2}\tanh\big(\frac{x}{2s_\textrm{h}}\big)$ of \citet{Stoiber2020b}, we obtain 
\begin{equation} \label{eq:aa11}
	\begin{split}
		p(d\mid r, \pmb{y}) \propto\
		&\bigg(\frac{1}{2} - \alpha_\textrm{h}\tanh\bigg(\frac{r-d}{2s_\textrm{h}}\bigg)\bigg)p_\textrm{f}(r)\,+ \\ 
		&\bigg(\frac{1}{2} + \alpha_\textrm{h}\tanh\bigg(\frac{r-d}{2s_\textrm{h}}\bigg)\bigg)p_\textrm{b}(r).
	\end{split}
\end{equation}
This is the same probability function as the one derived in Sect.~\ref{ssec:p5}.
Note, however, that in Sect.~\ref{ssec:p5} the smoothed step functions $h_\textrm{f}$ and $h_\textrm{b}$ from Eq.~(\ref{eq:p50}) and (\ref{eq:p51}) were used instead of a noise model $m_\textrm{n}$ to take into account a defined constant uncertainty.
In conclusion, this shows that extending the probabilistic model with a noise model $m_\textrm{n}$ and using the foreground and background probabilities $p(m_\textrm{f})=p(m_\textrm{b})=\alpha_\textrm{h}$ is equivalent to the introduction of a simple amplitude parameter $\alpha_\textrm{h}$ into the smoothed step functions.

\section{Derivative of Log-Posterior}\label{sec:ab}
To analyze the posterior probability distribution, it is desirable to have a closed-form solution that allows an easy interpretation.
In the following, we will thus derive a general formulation for the first-order derivative of the log-posterior, which will then be used in Appendix~\ref{sec:ac} to calculate the posterior probability distribution for specific parameter configurations.
Note that the derivation is similar to the proof of Gaussian equivalence developed in our previous work \citep{Stoiber2020b}.

For the derivation, we assume a contour at the line center and perfect step functions for the pixel-wise posterior probabilities defined by
\begin{gather}\label{eq:ab0}
	p_\textrm{f}(r) = \frac{1}{2} - \frac{1}{2}\sgn(r),\\ \label{eq:ab0_1}
	p_\textrm{b}(r) = \frac{1}{2} + \frac{1}{2}\sgn(r).
\end{gather}
Also, we consider infinitesimally small pixels and write the posterior probability distribution from Eq.~\ref{eq:p60} in continuous form for an infinite correspondence line
\begin{equation}\label{eq:ab1}
	p(d\mid\omega,\pmb{l}) \propto \prod_{r=-\infty}^\infty \big(h_\textrm{f}(r-d)p_\textrm{f}(r) +h_\textrm{b}(r-d)p_\textrm{b}(r)\big)^{\d{r}}.
\end{equation}

Starting from those assumptions, we first convert the product integral to the classical Riemann integral
\begin{equation}\label{eq:ab2}
	\begin{split}
		p(d\mid\omega,\pmb{l}) \propto \exp\bigg(\int_{r=-\infty}^\infty \ln\big(
		&h_\textrm{f}(r-d)p_\textrm{f}(r)\,+\\
		&h_\textrm{b}(r-d)p_\textrm{b}(r)\big)\d{r}\bigg).
	\end{split}\hspace{-1cm}
\end{equation}
The integral is then split at $r=0$, and the pixel-wise posterior probabilities from Eq.~(\ref{eq:ab0}) and (\ref{eq:ab0_1}) are introduced
\begin{equation}\label{eq:ab3}
	\begin{split}
		p(d\mid\omega,\pmb{l}) \propto \exp\bigg(
		&\int_{r=-\infty}^0 \ln\big(h_\textrm{f}(r-d)\big)\d{r}\,+\\
		&\int_{r=0}^\infty \ln\big(h_\textrm{b}(r-d)\big)\d{r}\bigg).
	\end{split}
\end{equation}
Finally, we substitute $x=r-d$ to write
\begin{equation}\label{eq:ab4}
	\begin{split}
		p(d\mid\omega,\pmb{l}) \propto \exp\bigg(
		&\int_{r=-\infty}^{-d} \ln\big(h_\textrm{f}(x)\big)\d{x}\,+\\
		&\int_{r=-d}^\infty \ln\big(h_\textrm{b}(x)\big)\d{x}\bigg).
	\end{split}
\end{equation}

The first-order derivative with respect to $d$ of the log-posterior can now be calculated using Leibniz's rule for differentiation under the integral
\begin{equation}\label{eq:ab5}
	\begin{split}
		\frac{\partial \ln\big(p(d\mid\omega,\pmb{l})\big)}{\partial d} = -\ln\big(h_\textrm{f}(-d)\big) + \ln\big(h_\textrm{b}(-d)\big).
	\end{split}
\end{equation}
We then adopt the definitions of the smoothed step functions from Eq.~(\ref{eq:p50}) and (\ref{eq:p51}) to write
\begin{equation}\label{eq:ab6}
	\begin{split}
		\frac{\partial \ln\big(p(d\mid\omega,\pmb{l})\big)}{\partial d} = 
		-\,&\ln\bigg(\frac{1}{2} - \alpha_\textrm{h}\tanh\bigg(\frac{-d}{2s_\textrm{h}}\bigg)\bigg)\,+ \\
		&\ln\bigg(\frac{1}{2} + \alpha_\textrm{h}\tanh\bigg(\frac{-d}{2s_\textrm{h}}\bigg)\bigg).
	\end{split}
\end{equation}
Finally, using the inverse hyperbolic tangent
\begin{equation}
	2\tanh^{-1}(x) = -\ln\bigg(\frac{1}{2}-\frac{x}{2}\bigg) + \ln\bigg(\frac{1}{2}+\frac{x}{2}\bigg),
\end{equation}
one is able to write the following closed-form expression for the first-order derivative of the log-posterior
\begin{equation}\label{eq:ab7}
	\frac{\partial\ln\big(p(d\mid\omega,\pmb{l})\big)}{\partial d} = -2\tanh^{-1}\bigg(2 \alpha_\textrm{h} \tanh\bigg(\frac{d}{2s_\textrm{h}}\bigg)\bigg).
\end{equation}

\section{Closed-Form Posteriors}\label{sec:ac}
Building on Appendix~\ref{sec:ab}, we derive closed-form posterior probability distributions for the two edge cases with either the amplitude parameter $\alpha_\textrm{h} = \frac{1}{2}$ or the slope parameter $s_\textrm{h} \to 0$.
We thereby start from the closed-form first-order derivative of the log-posterior that is given in Eq.~(\ref{eq:ab7}).
The full distribution can then be calculated using integration with a subsequent normalization
\begin{equation}\label{eq:ac0}
	p(d\mid\omega,\pmb{l}) \propto \exp\bigg(\int\frac{\partial\ln\big(p(d\mid\omega,\pmb{l})\big)}{\partial d} \d{d}\bigg).
\end{equation}

For the case with an amplitude parameter $\alpha_\textrm{h} = \frac{1}{2}$, the first-order derivative in Eq.~(\ref{eq:ab7}) simplifies to
\begin{equation}\label{eq:ac1}
	\frac{\partial\ln\big(p(d\mid\omega,\pmb{l})\big)}{\partial d} = -\frac{d}{s_\textrm{h}}.
\end{equation}
Introducing this term in Eq.~(\ref{eq:ac0}) and calculating the integral leads to the following expression for the posterior probability distribution
\begin{equation}\label{eq:ac2}
	p(d\mid\omega,\pmb{l}) \propto \exp\bigg(-\frac{d^2}{2s_\textrm{h}}\bigg).
\end{equation}
Because the posterior probability distribution has to be a valid \ac{PDF} that integrates to one, there is only one possible solution.
Knowing that, except for a constant scaling factor, the function looks like a Gaussian distribution, the final solution can only be the Gaussian distribution itself
\begin{equation}\label{eq:ac3}
	p(d\mid\omega,\pmb{l}) = \frac{1}{\sqrt{2\pi s_\textrm{h}}}\exp\bigg(-\frac{d^2}{2s_\textrm{h}}\bigg).
\end{equation}

For configurations with a slope parameter $s_\textrm{h}\to0$, the first-order derivative in Eq.~(\ref{eq:ab7}) simplifies to
\begin{equation}\label{eq:ac4}
	\frac{\partial\ln\big(p(d\mid\omega,\pmb{l})\big)}{\partial d} = -2\tanh^{-1}(2 \alpha_\textrm{h})\sgn(d).
\end{equation}
Introducing this term in Eq.~(\ref{eq:ac0}) and calculating the integral leads to the following closed-form posterior probability distribution
\begin{equation}\label{eq:ac5}
	p(d\mid\omega,\pmb{l}) \propto \exp\big(-2\tanh^{-1}(2 \alpha_\textrm{h})|d|\big).
\end{equation}
Similar to the previous derivation, we know that, with the exception of a constant scaling factor, the function is equal to a Laplace distribution, which again is a valid \ac{PDF}.
Introducing the scale parameter $b$, the final solution can thus only be the Laplace distribution itself
\begin{equation}\label{eq:ac6}
	p(d\mid\omega,\pmb{l}) = \frac{1}{2b}\exp\bigg(-\frac{|d|}{b}\bigg),\quad b=\frac{1}{2\tanh^{-1}(2\alpha_\textrm{h})}.
\end{equation}

\section{Inverse-Variance Weighting}\label{sec:ad}
In the following, we demonstrate that the derivatives for the local optimization that were defined in Sect.~\ref{ssec:t4} can be derived using inverse-variance weighting and a constant curvature of $\frac{1}{\alpha_\textrm{s}}$ for the second-order derivative.
Instead of the joint posterior probability defined in Eq.~(\ref{eq:t22}), we start with an energy function that combines probabilities from individual correspondence lines using inverse-variance weighting
\begin{equation}\label{eq:ad0}
	E(\pmb{\theta})=\sum_{i=1}^{n_\textrm{c}}\frac{1}{\sigma_i^2}\ln\big(p(d_{\textrm{s}i}(\pmb{\theta})\mid\omega_{\textrm{s}i},\pmb{l}_{\textrm{s}i})\big).
\end{equation}
Based on this function, the gradient vector and the Hessian matrix are calculated as the first- and second-order derivative with respect to $\pmb{\theta}$
\begin{gather}\label{eq:ad1}
	\pmb{g}^\top = \sum_{i=1}^{n_\textrm{c}}\frac{1}{\sigma_i^2}
	\frac{\partial\ln\big(p(d_{\textrm{s}i}\mid\omega_{\textrm{s}i},\pmb{l}_{\textrm{s}i})\big)}{\partial d_{\textrm{s}i}}
	\frac{\partial d_{\textrm{s}i}}{\partial \pmb{\theta}}
	\bigg\vert_{\pmb{\theta}=\pmb{0}},\\\label{eq:ad1_0}
	\begin{split}
	\pmb{H} \approx \sum_{i=1}^{n_\textrm{c}}
	\frac{1}{\sigma_i^2}
	&\frac{\partial^2\ln\big(p(d_{\textrm{s}i}\mid\omega_{\textrm{s}i},\pmb{l}_{\textrm{s}i})\big)}{\partial {d_{\textrm{s}i}}^2}\\
	&\left(\frac{\partial d_{\textrm{s}i}}{\partial \pmb{\theta}}\right)^{\hspace{-2pt}\top}\hspace{-2pt}
	\left(\frac{\partial d_{\textrm{s}i}}{\partial \pmb{\theta}}\right)
	\bigg\vert_{\pmb{\theta}=\pmb{0}}.
	\end{split}
\end{gather}
For the first-order derivative of the scaled contour distance $d_{\textrm{s}i}$, the derivations from Eq.~(\ref{eq:t42}) and (\ref{eq:t43}) can be used.
In contrast to Eq.~(\ref{eq:t45}), we use the definition of finite differences without a weighting term to calculate the first-order derivative of the log-posterior
\begin{equation}\label{eq:ad3}
	\frac{\partial\ln\big(p(d_{\textrm{s}i}\mid\omega_{\textrm{s}i},\pmb{l}_{\textrm{s}i})\big)}{\partial {d_{\textrm{s}i}}} \approx \ln\bigg(\frac{p(d_{\textrm{s}i}^+\mid\omega_{\textrm{s}i},\pmb{l}_{\textrm{s}i})}{p(d_{\textrm{s}i}^-\mid\omega_{\textrm{s}i},\pmb{l}_{\textrm{s}i})}\bigg),
\end{equation}
where $d_{\textrm{s}i}^-$ and $d_{\textrm{s}i}^+$ are again the two discrete contour distances that are closest to $d_{\textrm{s}i}(\pmb{\theta})$.
Because the variance is already considered in the energy function, we simply define a constant curvature for the second-order derivative of the log-posterior
\begin{equation}\label{eq:ad4}
	\frac{\partial^2\ln\big(p(d_{\textrm{s}i}\mid\omega_{\textrm{s}i},\pmb{l}_{\textrm{s}i})\big)}{\partial {d_{\textrm{s}i}}^2} \approx \frac{1}{\alpha_\textrm{s}}.
\end{equation}
Knowing that constant scaling terms do not affect the Newton optimization, both the gradient vector and the Hessian matrix can be multiplied with the step size $\alpha_\textrm{s}$.
Together with the inverse variance $\frac{1}{\sigma_i^2}$ that is already present in Eq.~(\ref{eq:ad1}) and (\ref{eq:ad1_0}), this results in exactly the same expressions for the gradient vector and the Hessian matrix as defined in Sect.~\ref{ssec:t4}.
In conclusion, the derivation thus shows that weighting the first-order derivative in Eq.~(\ref{eq:t45}) with a factor $\frac{\alpha_\textrm{s}}{\sigma_i^2}$ is the same as using inverse-variance weighting and a constant curvature of $\frac{1}{\alpha_\textrm{s}}$ for the second-order derivative.

\end{appendices}


%
%

\bibliographystyle{spbasic}      
\bibliography{09_literature}   

\end{document}